\documentclass[10pt,twocolumn,letterpaper]{article}

\usepackage[pagenumbers]{cvpr} 

\usepackage[dvipsnames]{xcolor}

\definecolor{cvprblue}{rgb}{0.21,0.49,0.74}
\usepackage[pagebackref,breaklinks,colorlinks,citecolor=cvprblue]{hyperref}

\title{DevilSight: Augmenting Monocular Human Avatar Reconstruction through a Virtual Perspective}

\author{ Yushuo Chen\textsuperscript{1} \qquad Ruizhi Shao\textsuperscript{1} \qquad Youxin Pang\textsuperscript{1} \qquad Hongwen Zhang\textsuperscript{2} \\ Xinyi Wu\textsuperscript{3} \qquad Rihui Wu\textsuperscript{3} \qquad Yebin Liu\textsuperscript{1}\\
\textsuperscript{1}Tsinghua University \qquad
\textsuperscript{2}Beijing Normal University \qquad
\textsuperscript{3}Honor Device Co., Ltd
}

\usepackage{multirow}

\begin{document}
\twocolumn[{%
\renewcommand\twocolumn[1][]{#1}%
\maketitle
\begin{center}
    \centering
    \captionsetup{type=figure}
    \includegraphics[width=\textwidth]{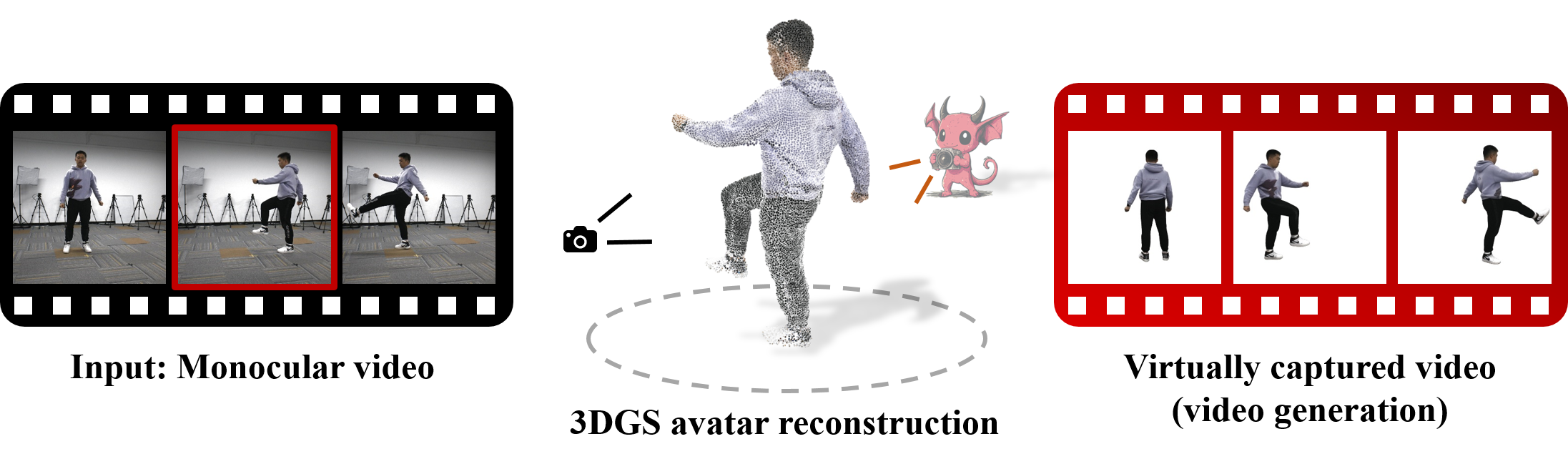}
    \captionof{figure}{We present DevilSight, which reconstruct 3DGS human avatar with fine-grained dynamic details from monocular video. Our approach utilizes a video generative model to ``see in the darkness'', generating a human video from alternative perspective. This enables us to capture high-frequency details from the input view while mitigating potential artifacts in unseen regions.}
\end{center}%
}]
\begin{abstract}
We present a novel framework to reconstruct human avatars from monocular videos.
Recent approaches have struggled either to capture the fine-grained dynamic details from the input or to generate plausible details at novel viewpoints, which mainly stem from the limited representational capacity of the avatar model and insufficient observational data.
To overcome these challenges, we propose to leverage the advanced video generative model, Human4DiT, to generate the human motions from alternative perspective as an additional supervision signal. 
This approach not only enriches the details in previously unseen regions but also effectively regularizes the avatar representation to mitigate artifacts.
Furthermore, we introduce two complementary strategies to enhance video generation: To ensure consistent reproduction of human motion, we inject the physical identity into the model through video fine-tuning. For higher-resolution outputs with finer details, a patch-based denoising algorithm is employed. 
Experimental results demonstrate that our method outperforms recent state-of-the-art approaches and validate the effectiveness of our proposed strategies.
\end{abstract}    
\section{Introduction}

Reconstructing high-fidelity and dynamically detailed human avatars holds significant value across a wide range of applications, including gaming, VR/AR technologies, and the movie industry.
However, despite notable advancements in human avatar reconstruction methods~\cite{liu2021neural,peng2021animatable,wang2022arah,li2023posevocab,zheng2022structured,zheng2023avatarrex,li2024animatable,chen2024meshavatar} utilizing multi-view inputs, it still remains a challenge to create an avatar with high-realistic texture details and dynamic fidelity from monocular video.

Previous approaches represent the human avatar with respect to deformable human templates~\cite{SMPL:2015, SMPL-X:2019}, in terms of structured latent codes~\cite{peng2021neural} or implicit 3D representations~\cite{weng2022humannerf,hu2024gauhuman,hu2024gaussianavatar,guo2023vid2avatar,jiang2022neuman,hu2025expressive}.
By constructing avatars in a canonical pose, these approaches align multi-frame observations into a unified 3D space and leverage pose-conditioned neural networks to interpolate dynamic details through spatial smoothness priors. While this framework ensures multi-view consistency, it struggles to recover high-frequency details due to limited cross-view correlation in the input frames and the required fusion operation.
Furthermore, the strong reliance on the naked body template limits its applicability in scenarios involving loose clothing.

Recently, generative models~\cite{rombach2022high,esser2024scaling,xie2024showo} show the potential to ``imagine'' previously unseen content from text or images. 
Moreover, some works~\cite{liu2023zero, voleti2024sv3d,long2024wonder3d} achieve the generation and reconstruction of 3D objects by adding camera control.
Based on these generative models, many methods~\cite{kolotouros2023dreamhuman, jiang2023avatarcraft, mendiratta2023avatarstudio,cao2023guide3d,jiang2023mvhuman,ho2024sith} leverage generative priors to enhance 3D reconstruction by score distillation sampling (SDS)~\cite{poole2022dreamfusion} or explicitly generating multi-view images as pseudo ground truths.
While these methods have achieved high-quality 3D static scene reconstruction from sparse-view inputs, reconstructing dynamic sequences remains a significant challenge. This difficulty primarily arises from generating video while maintaining the identity consistency across time and viewpoints.
In this paper, we propose leveraging Human4DiT~\cite{shao2024human4dit} to generate an identical motion sequence from an alternative perspective, thereby providing supplementary supervision signals for reconstruction. Recent advancements have demonstrated the impressive capability of diffusion transformers (DiT)~\cite{peebles2023scalable} to incorporate control signals from various dimensions. Specifically designed to produce spatio-temporally coherent human videos, Human4DiT supplies the view and temporal consistency priors essential to our approach. 
In practice, we choose to generate only one video from the back view, for its efficiency and effectiveness to address the majority of unseen regions.

Nevertheless, directly generating the back-view video using Human4DiT presents challenges in accurately reproducing human motions that are physically consistent with the input video. This limitation arises because the generation is conditioned on an identity embedding extracted from a single reference image, which is insufficient to encapsulate the dynamic characteristics inherent to human identity.
Consequently, generative models pretrained exclusively on such embeddings also struggle to accurately reproduce the motion.
Drawing inspiration from DreamBooth~\cite{ruiz2023dreambooth}, we propose Physical Identity Inversion through model finetuning, enabling video generation that aligns with the physical motion present in the input video. To prevent overfitting and maintain view consistency priors, we have meticulously designed our fine-tuning strategy and carefully selected the parameters involved.
Moreover, in order to bridge the resolution gap between captured and generated videos and to mitigate potential artifacts arising from this difference~\cite{Yu2023MipSplatting}, we have developed an innovative algorithm that effectively doubles the resolution of generated videos.
Given that DiT-based models are constrained to denoising at fixed resolution, the algorithm involves generating videos by partitioning them into patches.
Benefiting from these technical designs, our method achieves superior reconstruction quality compared to previous state-of-the-art methods.

As shown in the experiments, our method achieves high-fidelity human reconstruction with dynamic details and supports dynamically consistent image synthesis at novel views.
In summary, our contributions are:
\begin{itemize}
    \item We propose a novel framework for high-fidelity dynamic human-centric reconstruction from monocular video, which leverage the capability of video generative model to complement the unseen perspective.
    \item We introduce Physical Identity Inversion through model fine-tuning, effectively embedding personalized visual and physical identity into the model, thereby enabling accurate generation of identical human motions from the rear-view perspective.
    \item We present a patch-based denoising strategy for super-resolution video generation, effectively bridging the resolution gap between input and generated videos. This approach enables the subsequent reconstructed avatar model to render images with consistent granularity across varying viewpoints.
\end{itemize}
\section{Related Work}

\subsection{Neural Rendering for Human Reconstruction}

Neural rendering, as it effectively bridging the 3D assets with 2D images, has emerged as a powerful technique for reconstructing human figures directly from images.
Many approaches served this problem as reconstructing the human avatar by leveraging shape priors~\cite{weng2022humannerf,wang2022arah,peng2021animatable,liu2021neural,peng2021neural} from parametric human body templates~\cite{SMPL:2015,SMPL-X:2019} or learning the skinning fields according to the skeletons~\cite{su2021nerf,li2022tava,su2022danbo,geng2023learning}.
The human dynamics is then decomposed into rigid motions driven by skeletons and non-rigid deformations predicted by a pose-conditioned neural network.
On the other hand, this strategy unifies appearance information from different frames into a common 3D space, facilitating avatar creation and synthesis under novel views and novel poses.
However, the quality of the reconstructed avatar is constrained by the ability to translate low-frequency pose parameters into high-frequency dynamic details.
PoseVocab~\cite{li2023posevocab} decomposed the pose latent vector into per-joint embeddings for richer pose encoding. SLRF~\cite{zheng2022structured,zheng2023avatarrex} divide the entire radiance field into smaller local radiance fields, enabling better representativeness for local body part.
Animatable Gaussians~\cite{li2024animatable} and MeshAvatar~\cite{chen2024meshavatar} represented the pose parameters as the corresponding SMPL position map and employed the powerful 2D networks to encode it, achieving the SOTA in reconstructing the details.

Due to the high costs associated with studio-based multi-view data capture, several studies~\cite{yu2023monohuman,jiang2022selfrecon,su2021nerf,peng2022selfnerf} have explored reconstructing humans from monocular images or videos.
Single-view scenarios are particularly challenging because inevitable errors in estimated poses can degrade 3D correspondences across frames. To address this issue, some research has focused on error correction networks~\cite{weng2022humannerf,jiang2022neuman}, while others have utilized trainable pose parameters~\cite{guo2023vid2avatar,jiang2023instantavatar}. Recently, 3DGS is incorporated to further improve the training and inference efficiency~\cite{hu2024gaussianavatar,hu2024gauhuman,wen2024gomavatar,hu2025expressive,peng2025rmavatar,shao2024splattingavatar, pang2024ash,kocabas2024hugs,lei2024gart,qian20243dgs,zielonka2023drivable,moon2024expressive}.
Nevertheless, in order to stabilize the rendering quality in novel view, which is invisible from input, these methods usually constrained the complexity of the pose-conditioned networks. This limitation may hinder the accurate reconstruction of dynamic details observed in the input images.
In contrast, our approach regularize the avatar representation using the priors from generative models, facilitating both reconstruction and novel view synthesis.

\begin{figure*}[t]
    \centering
    \includegraphics[width=\linewidth]{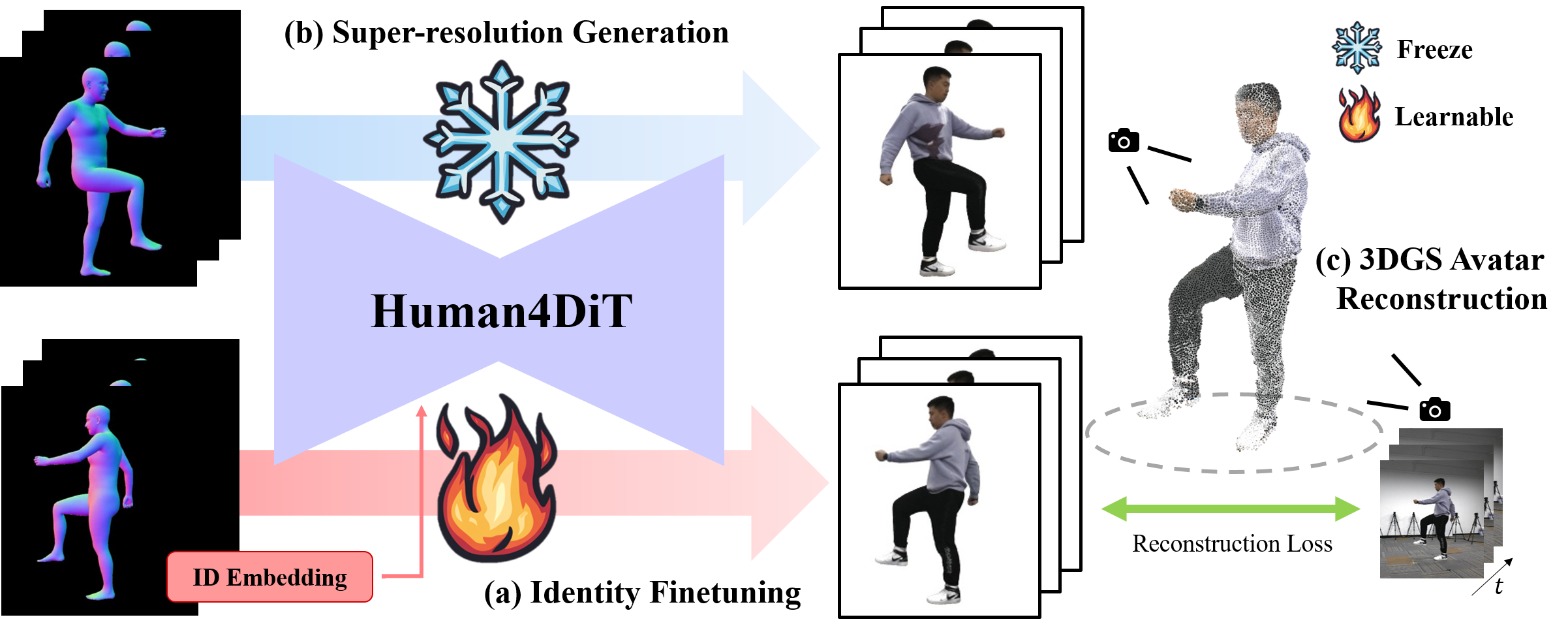}
    \caption{\textbf{Overview of our method.} Our method leverages priors from Human4DiT to enable robust monocular avatar reconstruction. It comprises three key components: (a) fine-tuning the model for personalized content generation, (b) generating consistent rear-view motion with superresolution, and (c) reconstructing 3DGS avatars using pseudo multi-view data.}
    \label{fig:overview}
\end{figure*}

\subsection{Generative Prior for Human Avatars}
With the development of generative models~\cite{rombach2022high,esser2024scaling,xie2024showo,chen2024videocrafter2,wang2024emu3,guo2024sparsectrl}, more and more methods~\cite{kolotouros2023dreamhuman, jiang2023avatarcraft, mendiratta2023avatarstudio,cao2023guide3d,jiang2023mvhuman,ho2024sith,wang2025wonderhuman,lee2024gtu} seek to inject human priors from large models into human avatar reconstruction.
They mainly leverage the power of generative models to iteratively optimize 3D representations.
For example, some methods employ score distillation sampling (SDS) loss~\cite{poole2022dreamfusion} to optimize 3D representations conditioned on text prompt~\cite{kolotouros2023dreamhuman, jiang2023avatarcraft, mendiratta2023avatarstudio,wang2024humancoser,xu2023seeavatar,huang2024humannorm,liao2024tada,huang2024tech,lee2024gtu}, skeletons~\cite{huang2023dreamwaltz, zeng2023avatarbooth, wang2024disentangled,liu2024humangaussian}, and densepose~\cite{zhang2024avatarverse}, normal maps. 
However, textual descriptions and 2D pose maps inherently lack precision in representing fine-grained geometric and texture details. 
Moreover, SDS primarily optimizes 3D parameters by enforcing distributional constraints, which can lead to issues such as oversaturation and excessive smoothing.
Although some methods~\cite{cao2024dreamavatar} alleviate the above problems by using variational score distillation (VSD) loss~\cite{wang2023prolificdreamer}, the computational cost is higher.

Due to the inherent limitations of SDS loss, subsequent works~\cite{cao2023guide3d,jiang2023mvhuman,ho2024sith,albahar2023single,chen2024generalizable} avoid using it and instead explicitly leverage generative models~\cite{rombach2022high,zhang2023adding,voleti2024sv3d}. 
These approaches typically generate multi-view images or videos using a pretrained large model, followed by direct 3D reconstruction and optimization. Human-related priors are learned from 2D structures and conditions (e.g., rendered skeletons or surface normals).
Similarly, HumanSplat~\cite{pan2025humansplat} generates multi-view features within latent space and reconstructs the human body through feature aggregation.
Recent approaches~\cite{liang2024diffusion4d,pan2024efficient4d,xie2024sv4d,yang2024diffusion,zhang20244diffusion} extends these strategies to video-to-4D generation. 
However, the absence of explicit 3D structural modeling exacerbates inherent temporal and view inconsistencies in 2D generative models, often leading to flat or blurry results. 
L4GM~\cite{ren2024l4gm} and GVFDiffusion~\cite{zhang2025gaussian} directly synthesize 4D content conditioned on the given video frames. Despite this, they primarily handle simple skinning-like motions and fail to capture complex dynamic details. Preserving the human identity while recovering fine-grained motions remains a significant challenge.


Our task formulation is similar to WonderHuman~\cite{wang2025wonderhuman}, as both leverage generative priors to reconstruct dynamic 3D human avatar from a monocular video.
WonderHuman optimize 3D human with SDS loss in both canonical and observation spaces to ensure visual consistency.
However, due to the oversaturation and excessive smoothing caused by SDS, the generated invisible regions lack fine details and exhibit noticeable inconsistencies.


\subsection{Identity Adaptation}
Maintaining identity consistency is crucial in digital humans, particularly when handling complex textures and dynamic movements.
Most methods~\cite{hu2024animate,xu2024magicanimate,wang2024disco,zhu2024champ} use a single reference image as input and design a reference network to embed the image into the backbone model. 
However, this approach demands a substantial amount of data for training and is computationally expensive.
In addition, IP-Adapter~\cite{ye2023ip} designs an effective and lightweight adapter to enable image prompt for pre-trained text-to-image diffusion models.
Specifically, IP-Adapter modifies the cross-attention mechanism by separating cross-attention layers for text and image features.
By freezing most parameters, image prompt can be learned by training only the image cross-attention layers.
DreamBooth~\cite{ruiz2023dreambooth} can fine-tune the generative model using a few reference images, ensuring identity consistency.
In details, DreamBooth fine-tunes a pretrained diffusion model to link a unique identifier to a specific subject, allowing it to be seamlessly embedded into different scenes.
VideoComposer~\cite{wang2023videocomposer} directly concatenates the reference image with noise to achieve identity injection.

\section{Method}

\subsection{Overview}

Given a monocular sequence of human motions with the corresponding human poses, which could be estimated using existing tools~\cite{goel2023humans,wang2024tram}, our task is to reconstruct a high-fidelity 3DGS avatar of this subject. 
High-frequency dynamic details often necessitate a trade-off between accurate input-view reconstruction and generalizable novel view synthesis. To address this, we propose leveraging advanced video generative models to generate additional videos from alternate perspectives as pseudo-supervision. This approach could effectively regularize the 3D representation from overfitting. 
However, generating a video that replicates human motions from an alternative viewpoint presents significant challenges. It requires preserving the individual's identity both visually and physically, ensuring that distinctive features and movement characteristics remain consistent across perspectives. 
Additionally, video generative models often face computational memory constraints that limit their output to low-resolution videos, typically below 1080p. It is crucial to bridge resolution gap between generated video and input video, to maintain visual fidelity and meet quality expectations. 
We propose physical identity inversion (Sec. \ref{sec:finetune}) through model finetuning and super-resolution generation (Sec. \ref{sec:sr}) to tackle these challenges, respectively. 
An overview of our method is presented in Fig. \ref{fig:overview}.

\begin{figure*}[t]
    \centering
    \includegraphics[width=\linewidth]{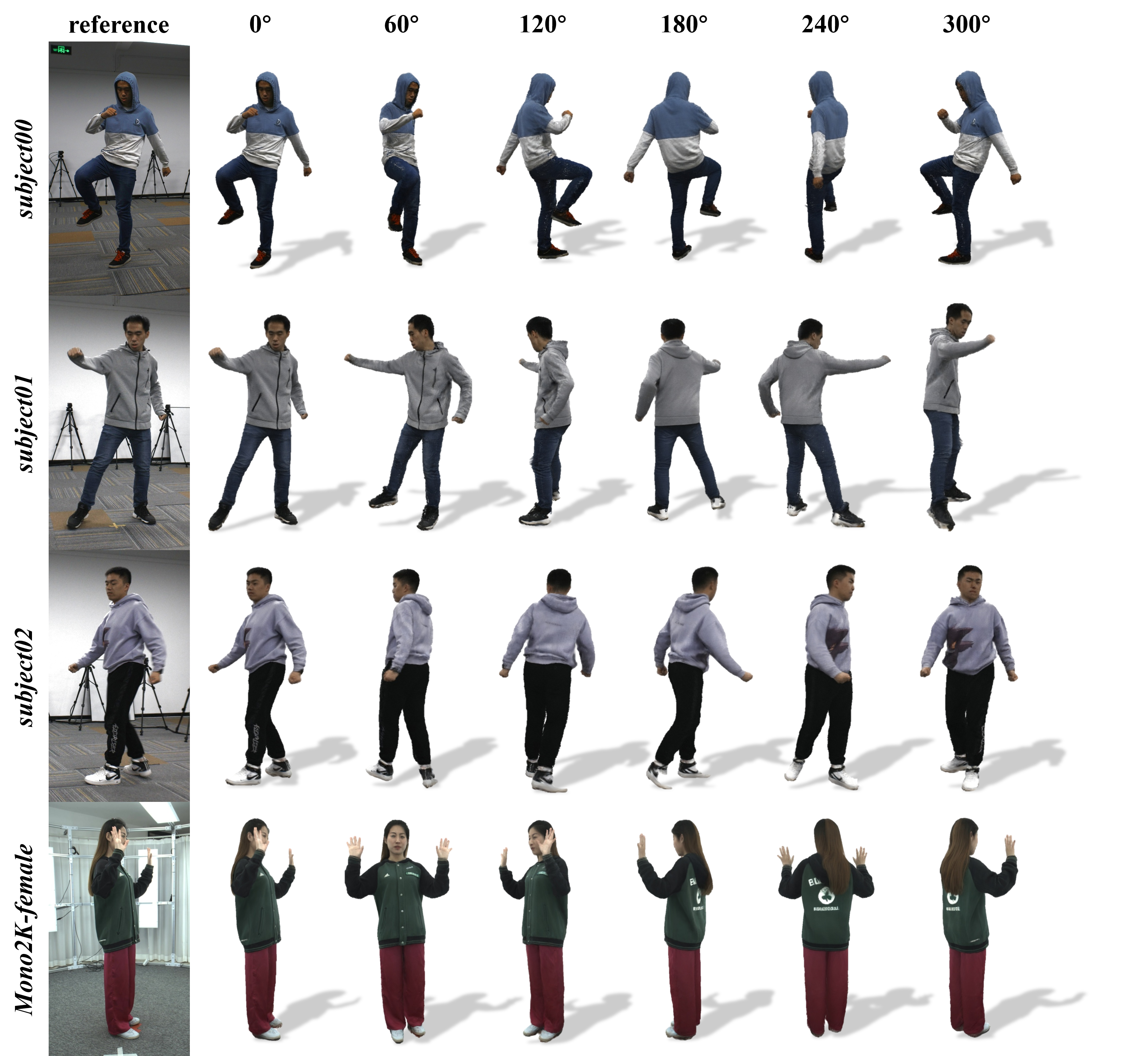}
    \caption{\textbf{Freeview Rendering of our reconstructed subjects from \textit{THuman4.0} and \textit{Mono2K}.} By leveraging the priors from Human4DiT, we not only accurately reconstruct dynamic details from the monocular input but also supports novel view synthesis with equivalent quality. Zoom in to see more details.}
    \label{fig:freeview}
\end{figure*}

\subsection{Preliminaries}

\textbf{Expressive 3DGS Avatar Representation.} 3D Gaussian Splatting~\cite{kerbl3Dgaussians} is a 3D point-based representation for efficient and realistic rendering. It's represented by a set of 3D Gaussians, each of which is parameterized by its 3D center position $\boldsymbol{\mu}$, a covariance matrix $\boldsymbol\Sigma$, opacity $\alpha$ and color $\boldsymbol{c}$, and distributed as:
\begin{equation}
    f(\mathbf{x}|\boldsymbol{\mu}, \boldsymbol{\Sigma}) \propto \exp\left( -\frac{1}{2}(\mathbf{x} - \boldsymbol{\mu})^\top \boldsymbol{\Sigma}^{-1}(\mathbf{x} - \boldsymbol{\mu})\right),
\end{equation}
where the covariance matrix is further parameterized by a rotation quaternion $\mathbf{q}$ and a 3D scaling vector $\mathbf{s}$ practically.

The rendering is conducted by splatting these 3D Gaussians onto the 2D plane, where the color of each pixel is computed by composition of the Gaussians overlapping on it. Following recent advancements on human avatars~\cite{li2024animatable,hu2024gaussianavatar}, we represent the avatar as 2D Gaussian map, and utilize UNet $\mathcal{U}$ to produce the Gaussian maps under different poses. Specifically, the Gaussian points are anchored onto the SMPL template by orthogonally projecting the canonical mesh to the front plane and back plane. Each active pixel corresponds to a 3D Gaussian point. 
Given any human pose $\boldsymbol{\Theta}=\{\theta_i\}_{i=1}^J$, these points could be deformed using the transformation matrix from LBS, and other Gaussian attributes are acquired by feeding the front-back projected posed position map~\cite{li2024animatable,chen2024meshavatar} $\mathcal P_f, \mathcal P_b$ to UNet:
\begin{equation}
    \mathcal{G}_f(\boldsymbol{\Theta}), \mathcal{G}_b(\boldsymbol{\Theta}) \leftarrow \mathcal{U}(\mathcal{P}_f(\boldsymbol{\Theta}), \mathcal{P}_b(\boldsymbol{\Theta})),
\end{equation}
where $\mathcal{G}=(\Delta\boldsymbol{\mu}, \mathbf{q}, \mathbf{s}, \alpha, \mathbf{c})$ consists of a pose-dependent point offset $\Delta\boldsymbol{\mu}$ to model non-rigid deformations.

\textbf{Human4DiT.}
Human4DiT~\cite{shao2024human4dit} is a DiT-based~\cite{peebles2023scalable} approach for generating high-quality, 360-degree human videos with spatio-temporal coherence from a single image.
By leveraging a hierarchical structure and the strength of DiT in capturing global features, it enables the synthesis of videos with strong 3D consistency.
Specifically, Human4DiT proposes a hierarchical 4D transformer architecture that factorizes self-attention across views, time steps, and spatial dimensions.
To enable precise control, dedicated modules are designed to embed camera parameters, temporal information, human motion, and identity.
Additionally, Human4DiT collects large-scale multi-dimensional datasets, including 2D videos, multi-view videos, 3D data, and 4D data. 
By employing a multi-dimensional training strategy, the model effectively leverages these diverse data to enhance the generalization ability.

Human4DiT iteratively denoises a gaussian noise $\epsilon$ to obtain a clean latent representation $z$, which is then passed through a decoder to generate the final output.
During training, the model is trained to predict the applied noise from the noisy latent $z_{t}$:
\begin{equation}
    \mathcal{L}_{\mathrm{diffusion}}=\mathbb{E}_{\mathbf{z}_t, c_{\mathrm{ref}}, c_{\Theta}, \epsilon, t}\left(\left\|\epsilon-\epsilon_\theta\left(\mathbf{z}_t, c_{\mathrm{ref}}, c_{\Theta}, t\right)\right\|_2^2\right) \label{eq:diffusion}
\end{equation}
where $\epsilon_\theta$ represents the denoising transformer, $t$ represents the timesteps, and $c_{\mathrm{ref}}$ and $c_{\Theta}$ represent the conditions of identity reference and human poses, respectively.

\begin{table*}[t]
\caption{\textbf{Quantitative comparisons} on input view reconstruction and novel view synthesis with HumanNeRF~\cite{weng2022humannerf}, GaussianAvatar~\cite{hu2024gaussianavatar} and AnimatableGaussians (AG for short)~\cite{li2024animatable}. The best and the second best are highlighted in \textbf{bold} and \underline{underlined} fonts, respectively.}
\label{tab-main}
\centering
\begin{tabular}{@{}ccccccccccc@{}}
\toprule
\multirow{2}{*}{Dataset}   & \multirow{2}{*}{Method}  & \multicolumn{3}{c}{Input View} & \multicolumn{3}{c}{Novel View 1} & \multicolumn{3}{c}{Novel View 2} \\ \cmidrule(lr){3-5} \cmidrule(lr){6-8} \cmidrule(l){9-11} 
                           &                & PSNR$\uparrow$     & SSIM$\uparrow$     & LPIPS$\downarrow$    & PSNR$\uparrow$      & SSIM$\uparrow$      & LPIPS$\downarrow$    & PSNR$\uparrow$      & SSIM$\uparrow$      & LPIPS$\downarrow$    \\ \midrule
\multirow{4}{*}{\rotatebox{90}{\small \textit{THuman4.0}}} & HumanNeRF      & $30.69$    & $0.9515$   & $0.0841$  & $25.84$    & $0.9438$    & $\underline{0.1051}$   & $24.97$    & $0.9372$    & $\underline{0.1163}$ \\
                           & GaussianAvatar & $31.13$    & $0.9709$   & $0.0820$    & $22.25$    & $0.9533$    & $0.1271$   & $22.02$    & $0.9500$      & $0.1346$    \\
                           & AG             & $\mathbf{34.06}$    & $\mathbf{0.9798}$   & $\mathbf{0.0554}$   & $\underline{26.55}$    & $\underline{0.9588}$    & $0.1071$   & $\underline{25.65}$    & $\underline{0.9542}$   & $0.1193$     \\
                           & Ours           & $\underline{32.97}$    & $\underline{0.9769}$   & $\underline{0.0558}$   & $\mathbf{26.79}$     & $\mathbf{0.9605}$    & $\mathbf{0.0995}$    & $\mathbf{25.98}$    & $\mathbf{0.9560}$     & $\mathbf{0.1111}$    \\ \midrule
\multirow{4}{*}{\rotatebox{90}{\small \textit{Mono2K}}}          & HumanNeRF      & $27.91$   & $0.9315$   & $0.0822$   & $26.02$    & $0.9305$    & $0.0873$    & $25.45$    & $0.9263$    & $\underline{0.0941}$   \\
                           & GaussianAvatar & $29.13$    & $0.9729$   & $0.0734$   & $21.26$    & $0.9549$    & $0.1071$   & $20.83$    & $0.9542$    & $0.1121$   \\
                           & AG             & $\mathbf{33.15}$    & $\mathbf{0.9807}$   & $\mathbf{0.0533}$   & $\mathbf{26.29}$    & $\underline{0.9654}$    & $\underline{0.0819}$    & $\underline{25.60}$    & $\underline{0.9591}$   & $0.0972$    \\
                           & Ours           & $\underline{32.19}$    & $\underline{0.9768}$   & $\underline{0.0614}$   & $\underline{26.26}$    & $\mathbf{0.9670}$    & $\mathbf{0.0788}$    & $\mathbf{25.73}$    & $\mathbf{0.9633}$     & $\mathbf{0.0898}$    \\ \bottomrule
\end{tabular}
\end{table*}

\subsection{Physical Identity Inversion by Finetuning} \label{sec:finetune}
In Human4DiT, the human identity is injected to the network through a CLIP~\cite{radford2021learning} embedding $c_{\mathrm{ref}}$ extracted from a reference image. This approach has several limitations.
On the one hand, the embedding conflates visual patterns from both the human subject and the background, leading to insufficient emphasis on the human identity. On the other hand, since the embedding is based on a single image, it provides an incomplete representation of the human identity, inadequately capturing the complex physical properties necessary for high-quality generation.

To overcome these limitations, we inject the human identity by fine-tuning the model using input video data. Specifically, we define a unique, learnable embedding, $c_{id}$, initialized from the CLIP embedding corresponding to the human subject, which enables a richer representation of identity beyond the constraints of a single-image reference.
The fine-tuning is performed using the reconstruction loss analogous to Equation \ref{eq:diffusion}. Additionally, to preserve capability to generate multi-view consistent video, we alternate tuning the model between the input video and the multi-view dataset from Human4DiT. Notably, only the attention layers conditioned on identity are fine-tuned.

\subsection{Back-view Generation with Super Resolution} \label{sec:sr}

In order to fully capture previously unseen details, we opted to generate the video from a rear perspective. In particular, based on the SMPL parameters of the first frame, we derived the new camera extrinsic matrix by rotating the original camera 180 degree around the axis passing through the root position and directed along the global orientation.

\begin{figure}[h]
    \centering
    \includegraphics[width=\linewidth]{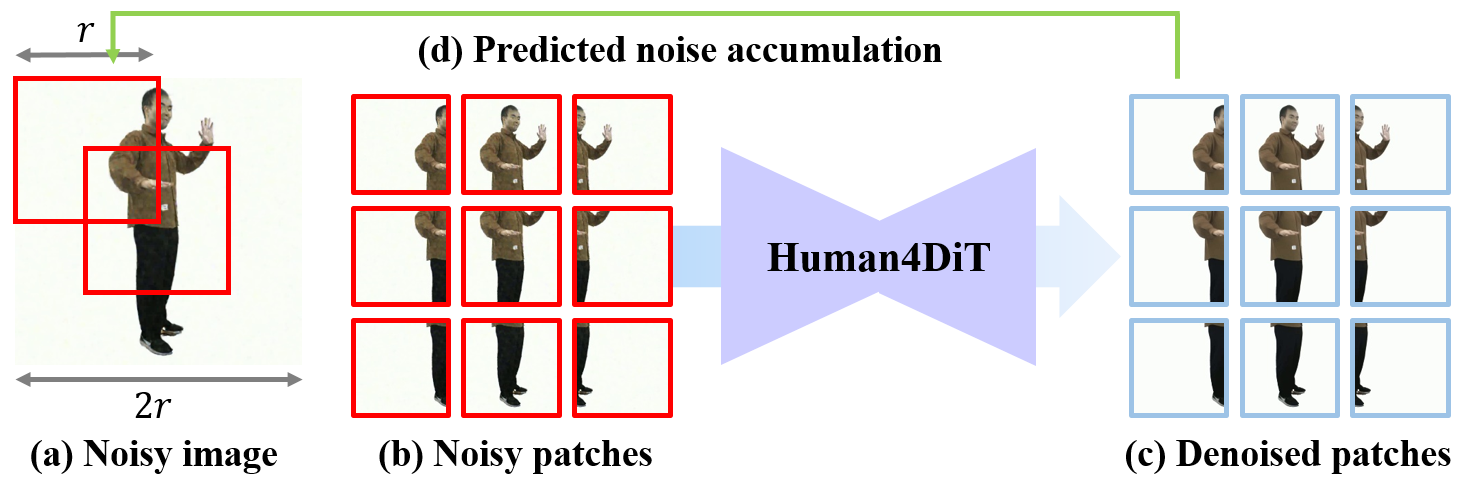}
    \caption{\textbf{Illustration of Super-resolution Generation.} At each diffusion timestep, we partition the image into 9 overlapping patches. The noises are then predicted independently by patch and accumulated by weighted sum. }
    \label{fig:method-sr}
\end{figure}

Human4DiT generates images at a standard resolution of $768\times 768$, which is considerably lower than the typical 2K resolution or higher found in most images. This discrepancy creates a pronounced resolution gap when transitioning from the front view to the back view. To address this, we instead generate images at twice the original resolution in both width and height.
Enlarging the canvas won't affect the decoding because of the shift invariance from VAE decoder. Consequently, we propose an algorithm that produces larger latent representations using a smaller, fixed-size diffusion denoising approach.
Specifically, we employ a similar sliding window strategy to maintain spatial consistency in the generated images, as how other methods preserve temporal consistency in video generation.
As illustrated in Fig. \ref{fig:method-sr}, at each denoising timestep the latent image is partitioned into nine overlapping patches; noise is predicted independently for each patch and then merged via a weighted sum. 
Correspondingly, during model fine-tuning (Sec. \ref{sec:finetune}), we incorporate randomly cropped videos into the training dataset to enhance patch-based generation.

\begin{figure*}[t]
    \centering
    \includegraphics[width=\linewidth]{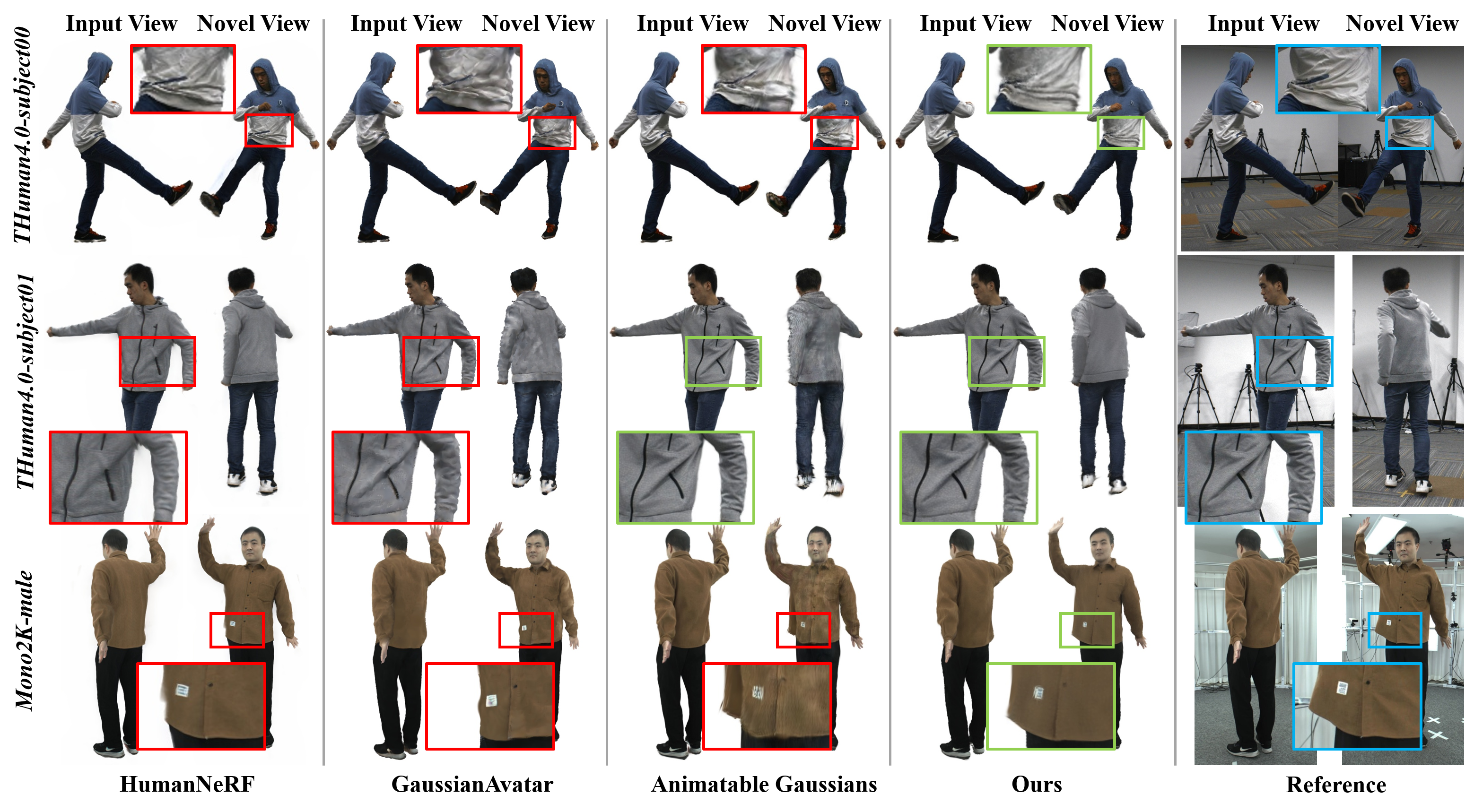}
    \caption{\textbf{Qualitative Comparisons} on input view reconstruction and novel view synthesis. Our approach achieves high-fidelity reconstruction by leveraging priors from Human4DiT. Zoom in to see more details.}
    \label{fig:compare}
\end{figure*}

After generating the back-view video, we directly treat it as a real captured video for pseudo supervision. 
\section{Experiments}

\textbf{Dataset.} 
We evaluated our method using the THuman4.0 dataset~\cite{zheng2022structured}, a multi-view dataset with a resolution of $1330\times 1150$, featuring characters with rich textures and dynamic details. To assess our method on higher-resolution images, we collected an additional dataset, named \textit{Mono2K}, comprising images at the resolution of $1500\times 2048$. For evaluation, we selected all three sequences (\textit{subject00}, \textit{subject01} and \textit{subject02}) from THuman4.0 and two typical sequences (\textit{Mono2K-male} and \textit{Mono2K-female}) from \textit{Mono2K}. Each sequence includes manually selected video clips featuring turning motions to ensure full-body visibility. For fair comparison in novel view synthesis, we utilized ground truth SMPL-X poses fitted from multiple views as input. For methods requiring SMPL poses, we converted them manually using the official tool~\cite{SMPL-X:2019}. Foreground masks were obtained using Segment Anything 2~\cite{ravi2024sam2}.

\textbf{Baselines.} 
We compared our method with several state-of-the-art approaches, including HumanNeRF~\cite{weng2022humannerf}, GaussianAvatar~\cite{hu2024gaussianavatar}, and Animatable Gaussians~\cite{li2024animatable}. Both GaussianAvatar and Animatable Gaussians are 3DGS-based methods for constructing human avatars, aiming to reconstruct dynamic details through 2D UNet architectures. GaussianAvatar is designed for monocular inputs, while Animatable Gaussians is tailored for multi-view inputs.
Due to the training efficiency of NeRF-based methods, when evaluating HumanNeRF on the \textit{Mono2K dataset}, we resized the images to $750\times 1024$. 

The comparison is also conducted with state-of-the-art video-to-4D approaches, L4GM~\cite{ren2024l4gm} and GVFDiffusion~\cite{zhang2025gaussian}, both of which are designed for general objects. L4GM leverages the generated multi-view videos, whereas GVFDiffusion adopts a holistic encode–decode paradigm to directly generate 4D content. All methods are reproduced using their publicly available codebases.

\textbf{Metrics.} We conducted our evaluation using established image quality metrics: Peak Signal-to-Noise Ratio (PSNR), Structure
Similarity Index Measure (SSIM)~\cite{wang2004image} and Learned Perceptual Image Patch Similarity (LPIPS)~\cite{zhang2018unreasonable}.
PSNR and SSIM are calculated over the entire image, with backgrounds set to white, while LPIPS is computed within the minimal square bounding box compassing the body. 

\subsection{Main Results}

Fig. \ref{fig:freeview} showcases our reconstruction results by rendering the reconstructed subject from different angles of viewpoints. Our approach demonstrates the capability not only in recovering the fine-grained details from the input view, but also produce plausible rendering quality at novel views. These results prove the robustness and adaptability of our approach in complementing missing details and regularizing dynamic human representations.

\subsection{Comparisons}

We assess the effectiveness of our approach through evaluations on both input view reconstruction and novel view synthesis.
The quantitative results are presented in Tab. \ref{tab-main}. As illustrated in Fig. \ref{fig:compare}, by leveraging priors from a video generative model, our method significantly outperforms other state-of-the-art approaches.
Due to its NeRF-based representation, HumanNeRF cannot accurately recover the fine details from the input, primarily producing dynamics through skinning.
In contrast, both GaussianAvatar and Animatable Gaussians demonstrate the superior representational capabilities of 3DGS for precise reconstruction. However, they exhibit artifacts in unseen regions, likely due to the lack of effective regularization for 3DGS, even though GaussianAvatar was specifically designed for monocular input.
Complementing the back view not only supplies additional reconstruction details but also regularizes the avatar representation, mitigating potential artifacts and enabling our method to achieve high-fidelity and high-quality reconstruction.


Moreover, as illustrated in the last example in Fig. \ref{fig:compare}, both HumanNeRF and GaussianAvatar are unable to accurately capture certain dynamic phenomena, such as the fluttering clothes, because their representations are tightly constrained to the human template.
Our approach leverages video generation that rigorously preserves the subject’s physical identity while ensuring view consistency with the input, enabling the reconstruction of these intricate dynamic details. This capability distinguishes our method from prior approaches, which have struggled to achieve similar results.

\begin{figure}[t]
    \centering
    \includegraphics[width=\linewidth]{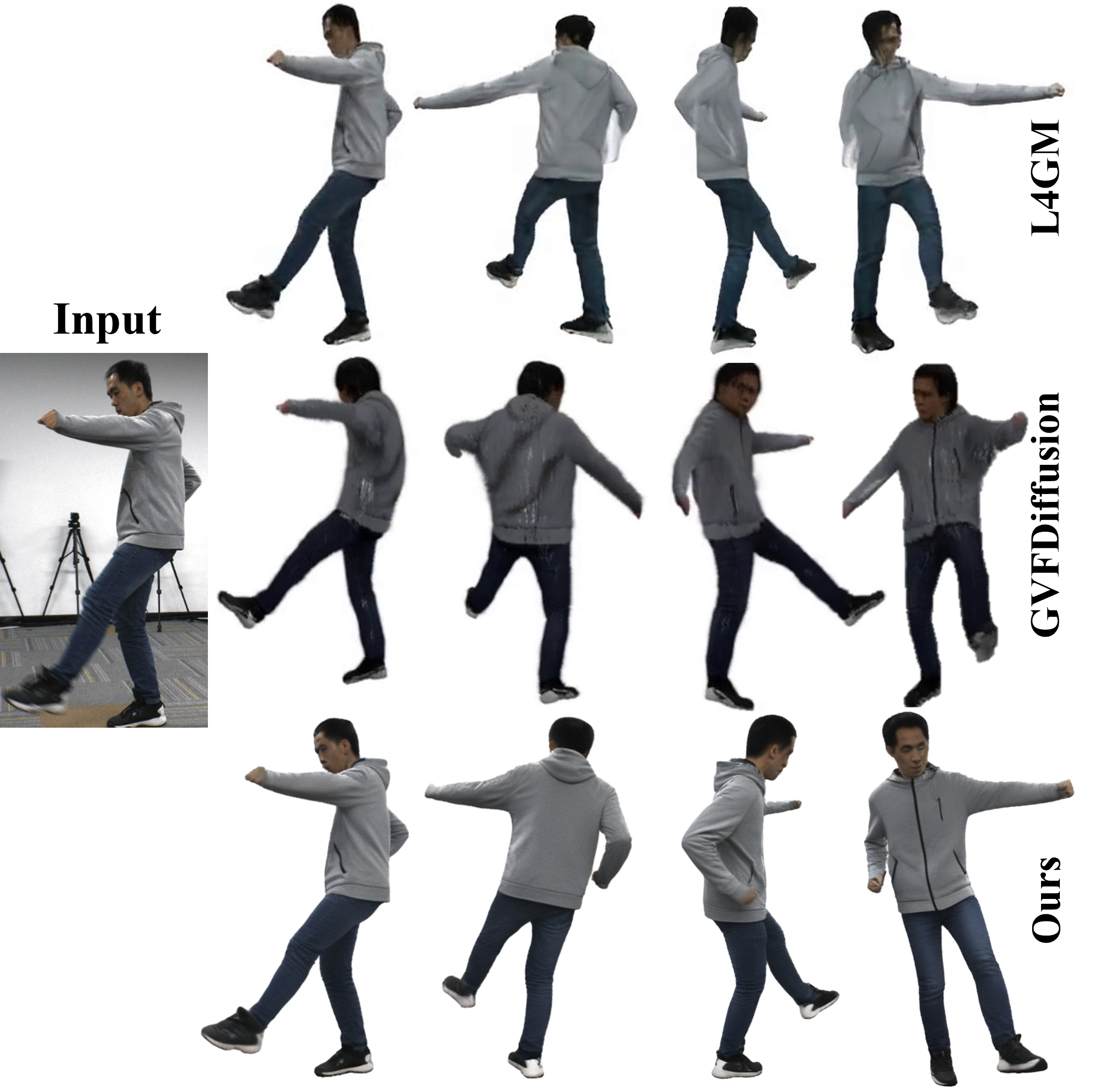}
    \caption{\textbf{Qualitative Comparisons} with SOTA video-to-4D methods. By leveraging both human priors and generative priors, our method effectively recovers fine-grained dynamic details.}
    \label{fig:4d}
\end{figure}

We further compare our method with SOTA video-to-4D approaches. As shown in Fig.~\ref{fig:4d}, both L4GM and GVFDiffusion struggle to reconstruct complex non-rigid dynamics present in the video. L4GM, due to the absence of human-body priors, fails to maintain consistent human identity across viewpoints and timestamps. GVFDiffusion, on the other hand, fails to encode the full range of motion details in the input video, leading to generated results that deviate substantially from the original content. In contrast, our approach benefits from both 3D human templates and generative priors, achieving high-fidelity dynamic reconstruction.

\subsection{Ablation Study}

\begin{figure}
    \centering
    \includegraphics[width=\linewidth]{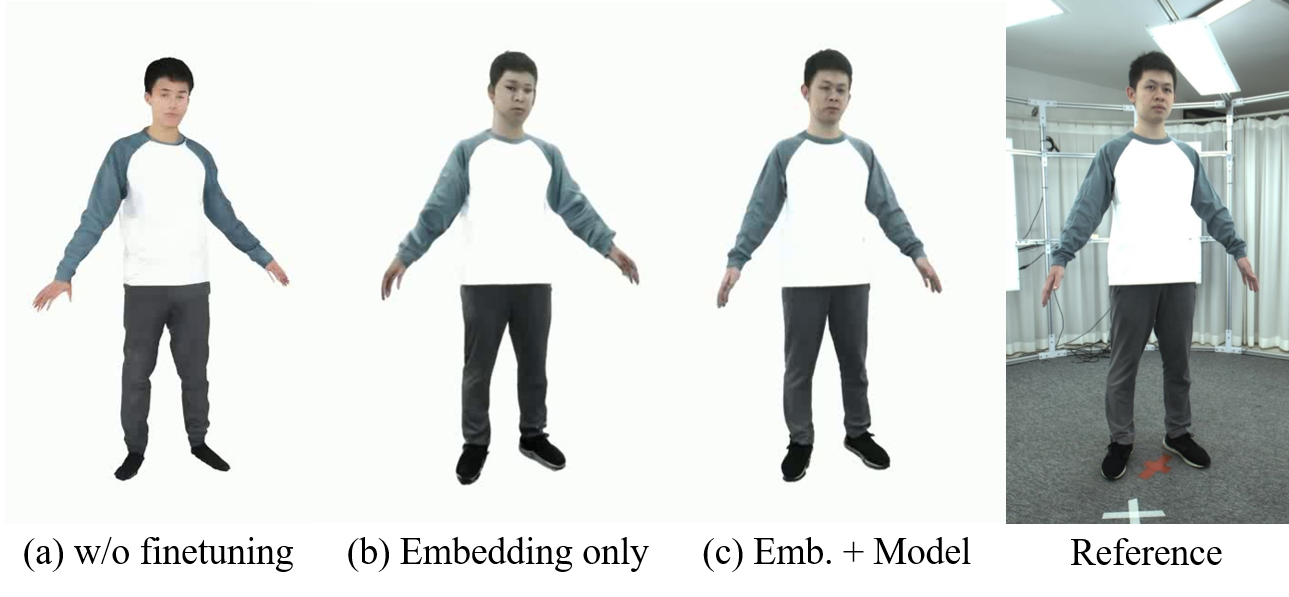}
    \caption{\textbf{Ablation study on different ID finetuning strategies.}}
    \label{fig:ablation-id}
\end{figure}

\textbf{ID Finetuning. } Fig. \ref{fig:ablation-id} illustrates the generation results of different finetuning strategies. 
(a) Without finetuning, Human4DiT generates identities with only approximate color similarity.
(b) Fine-tuning only the identity embedding yields a generally accurate appearance but lacks precision in details, such as clothing wrinkles, due to insufficient incorporation of the subject's dynamic characteristics.
(c) Finetuning along with the Human4DiT model implicitly injects the physical properties to generate dynamics. This comprehensive fine-tuning leads to the most accurate reconstruction of the human identity, enhancing the quality of subsequent back-view video generation.

\begin{figure}
    \centering
    \includegraphics[width=\linewidth]{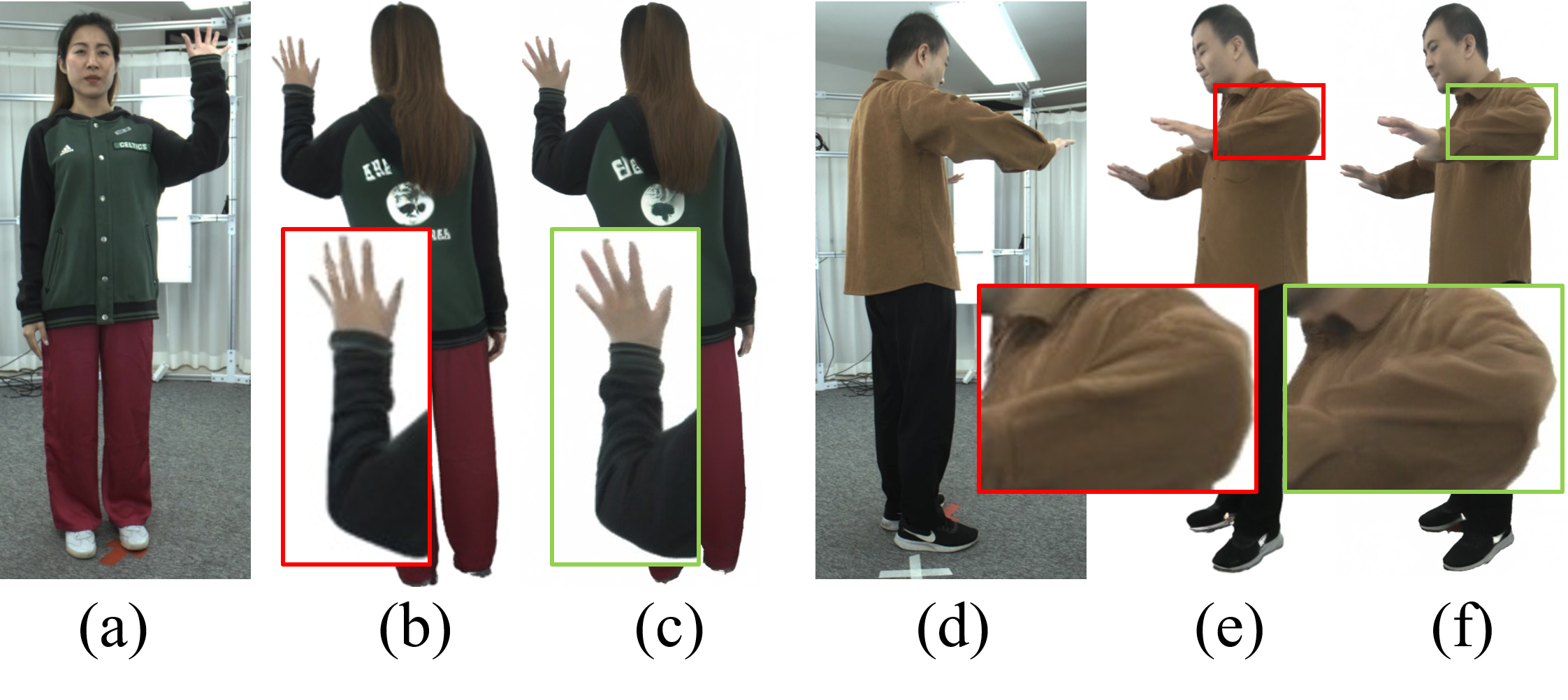}
    \caption{\textbf{Ablation study on super-resolution generation.} (a)(d) reference image from input view. (b)(e) standard-resolution generation. (c)(f) super-resolution generation. Zoom in to see details.}
    \label{fig:ablation-sr}
\end{figure}

\textbf{Super-resolution Generation.} 
Fig. \ref{fig:ablation-sr} presents the back-view generation results at different resolutions. As shown in (b) and (e), the absence of super-resolution leads to blurred clothing wrinkles, as diffusion models struggle to capture fine-grained local details. In contrast, our super-resolution approach not only enhances visual clarity and preserves intricate details but also generates high-resolution images that are crucial for improving the quality and robustness of avatar training.

\section{Conclusion}
We present a novel framework for high-fidelity dynamic human reconstruction from monocular video. By leveraging video generation to incorporate alternative viewpoints, our approach not only recovers high-frequency dynamic details from the input view but also supports novel view synthesis with equivalent quality. Furthermore, we enhance video generation by employing Physical Identity Inversion through model fine-tuning in conjunction with patch-based super-resolution techniques.

\textbf{Limitations.} Our approach is constrained by the current capabilities of the avatar model. As a pose-conditioned network, the avatar representation faces challenges in reconstructing data with pose-appearance one-to-many issue, a scenario commonly observed with loose clothing. 


{
    \small
    \bibliographystyle{ieeenat_fullname}
    \bibliography{main}

\begin{thebibliography}{93}
\providecommand{\natexlab}[1]{#1}
\providecommand{\url}[1]{\texttt{#1}}
\expandafter\ifx\csname urlstyle\endcsname\relax
  \providecommand{\doi}[1]{doi: #1}\else
  \providecommand{\doi}{doi: \begingroup \urlstyle{rm}\Url}\fi

\bibitem[AlBahar et~al.(2023)AlBahar, Saito, Tseng, Kim, Kopf, and Huang]{albahar2023single}
Badour AlBahar, Shunsuke Saito, Hung-Yu Tseng, Changil Kim, Johannes Kopf, and Jia-Bin Huang.
\newblock Single-image 3d human digitization with shape-guided diffusion.
\newblock In \emph{SIGGRAPH Asia 2023 Conference Papers}, pages 1--11, 2023.

\bibitem[Cao et~al.(2023)Cao, Cao, Han, Shan, and Wong]{cao2023guide3d}
Yukang Cao, Yan-Pei Cao, Kai Han, Ying Shan, and Kwan-Yee~K Wong.
\newblock Guide3d: Create 3d avatars from text and image guidance.
\newblock \emph{arXiv preprint arXiv:2308.09705}, 2023.

\bibitem[Cao et~al.(2024)Cao, Cao, Han, Shan, and Wong]{cao2024dreamavatar}
Yukang Cao, Yan-Pei Cao, Kai Han, Ying Shan, and Kwan-Yee~K Wong.
\newblock Dreamavatar: Text-and-shape guided 3d human avatar generation via diffusion models.
\newblock In \emph{Proceedings of the IEEE/CVF Conference on Computer Vision and Pattern Recognition}, pages 958--968, 2024.

\bibitem[Chen et~al.(2024{\natexlab{a}})Chen, Zhang, Cun, Xia, Wang, Weng, and Shan]{chen2024videocrafter2}
Haoxin Chen, Yong Zhang, Xiaodong Cun, Menghan Xia, Xintao Wang, Chao Weng, and Ying Shan.
\newblock Videocrafter2: Overcoming data limitations for high-quality video diffusion models, 2024{\natexlab{a}}.

\bibitem[Chen et~al.(2024{\natexlab{b}})Chen, Li, Zhang, Zhu, Huang, Chen, and Lee]{chen2024generalizable}
Jinnan Chen, Chen Li, Jianfeng Zhang, Lingting Zhu, Buzhen Huang, Hanlin Chen, and Gim~Hee Lee.
\newblock Generalizable human gaussians from single-view image.
\newblock \emph{arXiv preprint arXiv:2406.06050}, 2024{\natexlab{b}}.

\bibitem[Chen et~al.(2024{\natexlab{c}})Chen, Zheng, Li, Xu, and Liu]{chen2024meshavatar}
Yushuo Chen, Zerong Zheng, Zhe Li, Chao Xu, and Yebin Liu.
\newblock Meshavatar: Learning high-quality triangular human avatars from multi-view videos.
\newblock In \emph{European Conference on Computer Vision}, pages 250--269. Springer, 2024{\natexlab{c}}.

\bibitem[Esser et~al.(2024)Esser, Kulal, Blattmann, Entezari, M{\"u}ller, Saini, Levi, Lorenz, Sauer, Boesel, et~al.]{esser2024scaling}
Patrick Esser, Sumith Kulal, Andreas Blattmann, Rahim Entezari, Jonas M{\"u}ller, Harry Saini, Yam Levi, Dominik Lorenz, Axel Sauer, Frederic Boesel, et~al.
\newblock Scaling rectified flow transformers for high-resolution image synthesis.
\newblock In \emph{Forty-first international conference on machine learning}, 2024.

\bibitem[Geng et~al.(2023)Geng, Peng, Xu, Bao, and Zhou]{geng2023learning}
Chen Geng, Sida Peng, Zhen Xu, Hujun Bao, and Xiaowei Zhou.
\newblock Learning neural volumetric representations of dynamic humans in minutes.
\newblock In \emph{Proceedings of the IEEE/CVF Conference on Computer Vision and Pattern Recognition}, pages 8759--8770, 2023.

\bibitem[Goel et~al.(2023)Goel, Pavlakos, Rajasegaran, Kanazawa, and Malik]{goel2023humans}
Shubham Goel, Georgios Pavlakos, Jathushan Rajasegaran, Angjoo Kanazawa, and Jitendra Malik.
\newblock Humans in 4d: Reconstructing and tracking humans with transformers.
\newblock In \emph{Proceedings of the IEEE/CVF International Conference on Computer Vision}, pages 14783--14794, 2023.

\bibitem[Guo et~al.(2023)Guo, Jiang, Chen, Song, and Hilliges]{guo2023vid2avatar}
Chen Guo, Tianjian Jiang, Xu Chen, Jie Song, and Otmar Hilliges.
\newblock Vid2avatar: 3d avatar reconstruction from videos in the wild via self-supervised scene decomposition.
\newblock In \emph{Proceedings of the IEEE/CVF Conference on Computer Vision and Pattern Recognition}, pages 12858--12868, 2023.

\bibitem[Guo et~al.(2024)Guo, Yang, Rao, Agrawala, Lin, and Dai]{guo2024sparsectrl}
Yuwei Guo, Ceyuan Yang, Anyi Rao, Maneesh Agrawala, Dahua Lin, and Bo Dai.
\newblock Sparsectrl: Adding sparse controls to text-to-video diffusion models.
\newblock In \emph{European Conference on Computer Vision}, pages 330--348. Springer, 2024.

\bibitem[Heusel et~al.(2017)Heusel, Ramsauer, Unterthiner, Nessler, and Hochreiter]{heusel2017gans}
Martin Heusel, Hubert Ramsauer, Thomas Unterthiner, Bernhard Nessler, and Sepp Hochreiter.
\newblock Gans trained by a two time-scale update rule converge to a local nash equilibrium.
\newblock \emph{Advances in neural information processing systems}, 30, 2017.

\bibitem[Ho et~al.(2024)Ho, Song, Hilliges, et~al.]{ho2024sith}
I Ho, Jie Song, Otmar Hilliges, et~al.
\newblock Sith: Single-view textured human reconstruction with image-conditioned diffusion.
\newblock In \emph{Proceedings of the IEEE/CVF Conference on Computer Vision and Pattern Recognition}, pages 538--549, 2024.

\bibitem[Hu et~al.(2025)Hu, Fan, Wu, Xi, Lee, Pavlakos, Wang, et~al.]{hu2025expressive}
Hezhen Hu, Zhiwen Fan, Tianhao Wu, Yihan Xi, Seoyoung Lee, Georgios Pavlakos, Zhangyang Wang, et~al.
\newblock Expressive gaussian human avatars from monocular rgb video.
\newblock \emph{Advances in Neural Information Processing Systems}, 37:\penalty0 5646--5660, 2025.

\bibitem[Hu(2024)]{hu2024animate}
Li Hu.
\newblock Animate anyone: Consistent and controllable image-to-video synthesis for character animation.
\newblock In \emph{Proceedings of the IEEE/CVF Conference on Computer Vision and Pattern Recognition}, pages 8153--8163, 2024.

\bibitem[Hu et~al.(2024{\natexlab{a}})Hu, Zhang, Zhang, Zhou, Liu, Zhang, and Nie]{hu2024gaussianavatar}
Liangxiao Hu, Hongwen Zhang, Yuxiang Zhang, Boyao Zhou, Boning Liu, Shengping Zhang, and Liqiang Nie.
\newblock Gaussianavatar: Towards realistic human avatar modeling from a single video via animatable 3d gaussians.
\newblock In \emph{Proceedings of the IEEE/CVF conference on computer vision and pattern recognition}, pages 634--644, 2024{\natexlab{a}}.

\bibitem[Hu et~al.(2024{\natexlab{b}})Hu, Hu, and Liu]{hu2024gauhuman}
Shoukang Hu, Tao Hu, and Ziwei Liu.
\newblock Gauhuman: Articulated gaussian splatting from monocular human videos.
\newblock In \emph{Proceedings of the IEEE/CVF conference on computer vision and pattern recognition}, pages 20418--20431, 2024{\natexlab{b}}.

\bibitem[Huang et~al.(2024{\natexlab{a}})Huang, Shao, Zhang, Zhang, Feng, Liu, and Wang]{huang2024humannorm}
Xin Huang, Ruizhi Shao, Qi Zhang, Hongwen Zhang, Ying Feng, Yebin Liu, and Qing Wang.
\newblock Humannorm: Learning normal diffusion model for high-quality and realistic 3d human generation.
\newblock In \emph{Proceedings of the IEEE/CVF Conference on Computer Vision and Pattern Recognition}, pages 4568--4577, 2024{\natexlab{a}}.

\bibitem[Huang et~al.(2023)Huang, Wang, Zeng, Cao, Qi, Shi, Zha, and Zhang]{huang2023dreamwaltz}
Yukun Huang, Jianan Wang, Ailing Zeng, He Cao, Xianbiao Qi, Yukai Shi, Zheng-Jun Zha, and Lei Zhang.
\newblock Dreamwaltz: Make a scene with complex 3d animatable avatars.
\newblock \emph{Advances in Neural Information Processing Systems}, 36:\penalty0 4566--4584, 2023.

\bibitem[Huang et~al.(2024{\natexlab{b}})Huang, Yi, Xiu, Liao, Tang, Cai, and Thies]{huang2024tech}
Yangyi Huang, Hongwei Yi, Yuliang Xiu, Tingting Liao, Jiaxiang Tang, Deng Cai, and Justus Thies.
\newblock Tech: Text-guided reconstruction of lifelike clothed humans.
\newblock In \emph{2024 International Conference on 3D Vision (3DV)}, pages 1531--1542. IEEE, 2024{\natexlab{b}}.

\bibitem[Jiang et~al.(2022{\natexlab{a}})Jiang, Hong, Bao, and Zhang]{jiang2022selfrecon}
Boyi Jiang, Yang Hong, Hujun Bao, and Juyong Zhang.
\newblock Selfrecon: Self reconstruction your digital avatar from monocular video.
\newblock In \emph{Proceedings of the IEEE/CVF Conference on Computer Vision and Pattern Recognition}, pages 5605--5615, 2022{\natexlab{a}}.

\bibitem[Jiang et~al.(2023{\natexlab{a}})Jiang, Wang, Zhang, Chai, He, Chen, and Liao]{jiang2023avatarcraft}
Ruixiang Jiang, Can Wang, Jingbo Zhang, Menglei Chai, Mingming He, Dongdong Chen, and Jing Liao.
\newblock Avatarcraft: Transforming text into neural human avatars with parameterized shape and pose control.
\newblock In \emph{Proceedings of the IEEE/CVF International Conference on Computer Vision}, pages 14371--14382, 2023{\natexlab{a}}.

\bibitem[Jiang et~al.(2023{\natexlab{b}})Jiang, Luo, Jiang, Wang, Yu, and Xu]{jiang2023mvhuman}
Suyi Jiang, Haimin Luo, Haoran Jiang, Ziyu Wang, Jingyi Yu, and Lan Xu.
\newblock Mvhuman: tailoring 2d diffusion with multi-view sampling for realistic 3d human generation.
\newblock \emph{arXiv preprint arXiv:2312.10120}, 2023{\natexlab{b}}.

\bibitem[Jiang et~al.(2023{\natexlab{c}})Jiang, Chen, Song, and Hilliges]{jiang2023instantavatar}
Tianjian Jiang, Xu Chen, Jie Song, and Otmar Hilliges.
\newblock Instantavatar: Learning avatars from monocular video in 60 seconds.
\newblock In \emph{Proceedings of the IEEE/CVF Conference on Computer Vision and Pattern Recognition}, pages 16922--16932, 2023{\natexlab{c}}.

\bibitem[Jiang et~al.(2022{\natexlab{b}})Jiang, Yi, Samei, Tuzel, and Ranjan]{jiang2022neuman}
Wei Jiang, Kwang~Moo Yi, Golnoosh Samei, Oncel Tuzel, and Anurag Ranjan.
\newblock Neuman: Neural human radiance field from a single video.
\newblock In \emph{European Conference on Computer Vision}, pages 402--418. Springer, 2022{\natexlab{b}}.

\bibitem[Kerbl et~al.(2023)Kerbl, Kopanas, Leimk{\"u}hler, and Drettakis]{kerbl3Dgaussians}
Bernhard Kerbl, Georgios Kopanas, Thomas Leimk{\"u}hler, and George Drettakis.
\newblock 3d gaussian splatting for real-time radiance field rendering.
\newblock \emph{ACM Transactions on Graphics}, 42\penalty0 (4), 2023.

\bibitem[Kocabas et~al.(2024)Kocabas, Chang, Gabriel, Tuzel, and Ranjan]{kocabas2024hugs}
Muhammed Kocabas, Jen-Hao~Rick Chang, James Gabriel, Oncel Tuzel, and Anurag Ranjan.
\newblock Hugs: Human gaussian splats.
\newblock In \emph{Proceedings of the IEEE/CVF conference on computer vision and pattern recognition}, pages 505--515, 2024.

\bibitem[Kolotouros et~al.(2023)Kolotouros, Alldieck, Zanfir, Bazavan, Fieraru, and Sminchisescu]{kolotouros2023dreamhuman}
Nikos Kolotouros, Thiemo Alldieck, Andrei Zanfir, Eduard Bazavan, Mihai Fieraru, and Cristian Sminchisescu.
\newblock Dreamhuman: Animatable 3d avatars from text.
\newblock \emph{Advances in Neural Information Processing Systems}, 36:\penalty0 10516--10529, 2023.

\bibitem[Lee et~al.(2024)Lee, Kim, and Joo]{lee2024gtu}
Inhee Lee, Byungjun Kim, and Hanbyul Joo.
\newblock Guess the unseen: Dynamic 3d scene reconstruction from partial 2d glimpses.
\newblock 2024.

\bibitem[Lei et~al.(2024)Lei, Wang, Pavlakos, Liu, and Daniilidis]{lei2024gart}
Jiahui Lei, Yufu Wang, Georgios Pavlakos, Lingjie Liu, and Kostas Daniilidis.
\newblock Gart: Gaussian articulated template models.
\newblock In \emph{Proceedings of the IEEE/CVF conference on computer vision and pattern recognition}, pages 19876--19887, 2024.

\bibitem[Li et~al.(2022)Li, Tanke, Vo, Zollh{\"o}fer, Gall, Kanazawa, and Lassner]{li2022tava}
Ruilong Li, Julian Tanke, Minh Vo, Michael Zollh{\"o}fer, J{\"u}rgen Gall, Angjoo Kanazawa, and Christoph Lassner.
\newblock Tava: Template-free animatable volumetric actors.
\newblock In \emph{European Conference on Computer Vision}, pages 419--436. Springer, 2022.

\bibitem[Li et~al.(2023)Li, Zheng, Liu, Zhou, and Liu]{li2023posevocab}
Zhe Li, Zerong Zheng, Yuxiao Liu, Boyao Zhou, and Yebin Liu.
\newblock Posevocab: Learning joint-structured pose embeddings for human avatar modeling.
\newblock In \emph{ACM SIGGRAPH 2023 conference proceedings}, pages 1--11, 2023.

\bibitem[Li et~al.(2024)Li, Zheng, Wang, and Liu]{li2024animatable}
Zhe Li, Zerong Zheng, Lizhen Wang, and Yebin Liu.
\newblock Animatable gaussians: Learning pose-dependent gaussian maps for high-fidelity human avatar modeling.
\newblock In \emph{Proceedings of the IEEE/CVF conference on computer vision and pattern recognition}, pages 19711--19722, 2024.

\bibitem[Liang et~al.(2024)Liang, Yin, Xu, Liang, Wang, Plataniotis, Zhao, and Wei]{liang2024diffusion4d}
Hanwen Liang, Yuyang Yin, Dejia Xu, Hanxue Liang, Zhangyang Wang, Konstantinos~N Plataniotis, Yao Zhao, and Yunchao Wei.
\newblock Diffusion4d: Fast spatial-temporal consistent 4d generation via video diffusion models.
\newblock \emph{arXiv preprint arXiv:2405.16645}, 2024.

\bibitem[Liao et~al.(2024)Liao, Yi, Xiu, Tang, Huang, Thies, and Black]{liao2024tada}
Tingting Liao, Hongwei Yi, Yuliang Xiu, Jiaxiang Tang, Yangyi Huang, Justus Thies, and Michael~J Black.
\newblock Tada! text to animatable digital avatars.
\newblock In \emph{2024 International Conference on 3D Vision (3DV)}, pages 1508--1519. IEEE, 2024.

\bibitem[Liu et~al.(2021)Liu, Habermann, Rudnev, Sarkar, Gu, and Theobalt]{liu2021neural}
Lingjie Liu, Marc Habermann, Viktor Rudnev, Kripasindhu Sarkar, Jiatao Gu, and Christian Theobalt.
\newblock Neural actor: Neural free-view synthesis of human actors with pose control.
\newblock \emph{ACM transactions on graphics (TOG)}, 40\penalty0 (6):\penalty0 1--16, 2021.

\bibitem[Liu et~al.(2023)Liu, Wu, Van~Hoorick, Tokmakov, Zakharov, and Vondrick]{liu2023zero}
Ruoshi Liu, Rundi Wu, Basile Van~Hoorick, Pavel Tokmakov, Sergey Zakharov, and Carl Vondrick.
\newblock Zero-1-to-3: Zero-shot one image to 3d object.
\newblock In \emph{Proceedings of the IEEE/CVF international conference on computer vision}, pages 9298--9309, 2023.

\bibitem[Liu et~al.(2024)Liu, Zhan, Tang, Shan, Zeng, Lin, Liu, and Liu]{liu2024humangaussian}
Xian Liu, Xiaohang Zhan, Jiaxiang Tang, Ying Shan, Gang Zeng, Dahua Lin, Xihui Liu, and Ziwei Liu.
\newblock Humangaussian: Text-driven 3d human generation with gaussian splatting.
\newblock In \emph{Proceedings of the IEEE/CVF Conference on Computer Vision and Pattern Recognition}, pages 6646--6657, 2024.

\bibitem[Long et~al.(2024)Long, Guo, Lin, Liu, Dou, Liu, Ma, Zhang, Habermann, Theobalt, et~al.]{long2024wonder3d}
Xiaoxiao Long, Yuan-Chen Guo, Cheng Lin, Yuan Liu, Zhiyang Dou, Lingjie Liu, Yuexin Ma, Song-Hai Zhang, Marc Habermann, Christian Theobalt, et~al.
\newblock Wonder3d: Single image to 3d using cross-domain diffusion.
\newblock In \emph{Proceedings of the IEEE/CVF conference on computer vision and pattern recognition}, pages 9970--9980, 2024.

\bibitem[Loper et~al.(2015)Loper, Mahmood, Romero, Pons-Moll, and Black]{SMPL:2015}
Matthew Loper, Naureen Mahmood, Javier Romero, Gerard Pons-Moll, and Michael~J. Black.
\newblock {SMPL}: A skinned multi-person linear model.
\newblock \emph{ACM Trans. Graphics (Proc. SIGGRAPH Asia)}, 34\penalty0 (6):\penalty0 248:1--248:16, 2015.

\bibitem[Mendiratta et~al.(2023)Mendiratta, Pan, Elgharib, Teotia, Tewari, Golyanik, Kortylewski, and Theobalt]{mendiratta2023avatarstudio}
Mohit Mendiratta, Xingang Pan, Mohamed Elgharib, Kartik Teotia, Ayush Tewari, Vladislav Golyanik, Adam Kortylewski, and Christian Theobalt.
\newblock Avatarstudio: Text-driven editing of 3d dynamic human head avatars.
\newblock \emph{ACM Transactions on Graphics (ToG)}, 42\penalty0 (6):\penalty0 1--18, 2023.

\bibitem[Moon et~al.(2024)Moon, Shiratori, and Saito]{moon2024expressive}
Gyeongsik Moon, Takaaki Shiratori, and Shunsuke Saito.
\newblock Expressive whole-body 3d gaussian avatar.
\newblock In \emph{European Conference on Computer Vision}, pages 19--35. Springer, 2024.

\bibitem[Pan et~al.(2025)Pan, Su, Lin, Fan, Zhang, Li, Shen, Mu, and Liu]{pan2025humansplat}
Panwang Pan, Zhuo Su, Chenguo Lin, Zhen Fan, Yongjie Zhang, Zeming Li, Tingting Shen, Yadong Mu, and Yebin Liu.
\newblock Humansplat: Generalizable single-image human gaussian splatting with structure priors.
\newblock \emph{Advances in Neural Information Processing Systems}, 37:\penalty0 74383--74410, 2025.

\bibitem[Pan et~al.(2024)Pan, Yang, Zhu, and Zhang]{pan2024efficient4d}
Zijie Pan, Zeyu Yang, Xiatian Zhu, and Li Zhang.
\newblock Efficient4d: Fast dynamic 3d object generation from a single-view video.
\newblock \emph{arXiv preprint arXiv:2401.08742}, 2024.

\bibitem[Pang et~al.(2024)Pang, Zhu, Kortylewski, Theobalt, and Habermann]{pang2024ash}
Haokai Pang, Heming Zhu, Adam Kortylewski, Christian Theobalt, and Marc Habermann.
\newblock Ash: Animatable gaussian splats for efficient and photoreal human rendering.
\newblock In \emph{Proceedings of the IEEE/CVF Conference on Computer Vision and Pattern Recognition}, pages 1165--1175, 2024.

\bibitem[Pavlakos et~al.(2019)Pavlakos, Choutas, Ghorbani, Bolkart, Osman, Tzionas, and Black]{SMPL-X:2019}
Georgios Pavlakos, Vasileios Choutas, Nima Ghorbani, Timo Bolkart, Ahmed A.~A. Osman, Dimitrios Tzionas, and Michael~J. Black.
\newblock Expressive body capture: {3D} hands, face, and body from a single image.
\newblock In \emph{Proceedings IEEE Conf. on Computer Vision and Pattern Recognition (CVPR)}, pages 10975--10985, 2019.

\bibitem[Peebles and Xie(2023)]{peebles2023scalable}
William Peebles and Saining Xie.
\newblock Scalable diffusion models with transformers.
\newblock In \emph{Proceedings of the IEEE/CVF international conference on computer vision}, pages 4195--4205, 2023.

\bibitem[Peng et~al.(2022)Peng, Hu, Zhou, and Zhang]{peng2022selfnerf}
Bo Peng, Jun Hu, Jingtao Zhou, and Juyong Zhang.
\newblock Selfnerf: Fast training nerf for human from monocular self-rotating video.
\newblock \emph{arXiv preprint arXiv:2210.01651}, 2022.

\bibitem[Peng et~al.(2021{\natexlab{a}})Peng, Dong, Wang, Zhang, Shuai, Zhou, and Bao]{peng2021animatable}
Sida Peng, Junting Dong, Qianqian Wang, Shangzhan Zhang, Qing Shuai, Xiaowei Zhou, and Hujun Bao.
\newblock Animatable neural radiance fields for modeling dynamic human bodies.
\newblock In \emph{Proceedings of the IEEE/CVF International Conference on Computer Vision}, pages 14314--14323, 2021{\natexlab{a}}.

\bibitem[Peng et~al.(2021{\natexlab{b}})Peng, Zhang, Xu, Wang, Shuai, Bao, and Zhou]{peng2021neural}
Sida Peng, Yuanqing Zhang, Yinghao Xu, Qianqian Wang, Qing Shuai, Hujun Bao, and Xiaowei Zhou.
\newblock Neural body: Implicit neural representations with structured latent codes for novel view synthesis of dynamic humans.
\newblock In \emph{Proceedings of the IEEE/CVF conference on computer vision and pattern recognition}, pages 9054--9063, 2021{\natexlab{b}}.

\bibitem[Peng et~al.(2025)Peng, Xie, Wang, Guo, Chen, Yang, and Dong]{peng2025rmavatar}
Sen Peng, Weixing Xie, Zilong Wang, Xiaohu Guo, Zhonggui Chen, Baorong Yang, and Xiao Dong.
\newblock Rmavatar: Photorealistic human avatar reconstruction from monocular video based on rectified mesh-embedded gaussians.
\newblock \emph{arXiv preprint arXiv:2501.07104}, 2025.

\bibitem[Poole et~al.(2022)Poole, Jain, Barron, and Mildenhall]{poole2022dreamfusion}
Ben Poole, Ajay Jain, Jonathan~T Barron, and Ben Mildenhall.
\newblock Dreamfusion: Text-to-3d using 2d diffusion.
\newblock \emph{arXiv preprint arXiv:2209.14988}, 2022.

\bibitem[Qian et~al.(2024)Qian, Wang, Mihajlovic, Geiger, and Tang]{qian20243dgs}
Zhiyin Qian, Shaofei Wang, Marko Mihajlovic, Andreas Geiger, and Siyu Tang.
\newblock 3dgs-avatar: Animatable avatars via deformable 3d gaussian splatting.
\newblock In \emph{Proceedings of the IEEE/CVF conference on computer vision and pattern recognition}, pages 5020--5030, 2024.

\bibitem[Radford et~al.(2021)Radford, Kim, Hallacy, Ramesh, Goh, Agarwal, Sastry, Askell, Mishkin, Clark, et~al.]{radford2021learning}
Alec Radford, Jong~Wook Kim, Chris Hallacy, Aditya Ramesh, Gabriel Goh, Sandhini Agarwal, Girish Sastry, Amanda Askell, Pamela Mishkin, Jack Clark, et~al.
\newblock Learning transferable visual models from natural language supervision.
\newblock In \emph{International conference on machine learning}, pages 8748--8763. PmLR, 2021.

\bibitem[Ravi et~al.(2024)Ravi, Gabeur, Hu, Hu, Ryali, Ma, Khedr, {R{"a}dle}, Rolland, Gustafson, Mintun, Pan, Alwala, Carion, Wu, Girshick, {Doll{\'a}r}, and Feichtenhofer]{ravi2024sam2}
Nikhila Ravi, Valentin Gabeur, Yuan-Ting Hu, Ronghang Hu, Chaitanya Ryali, Tengyu Ma, Haitham Khedr, Roman {R{"a}dle}, Chloe Rolland, Laura Gustafson, Eric Mintun, Junting Pan, Kalyan~Vasudev Alwala, Nicolas Carion, Chao-Yuan Wu, Ross Girshick, Piotr {Doll{\'a}r}, and Christoph Feichtenhofer.
\newblock Sam 2: Segment anything in images and videos.
\newblock \emph{arXiv preprint arXiv:2408.00714}, 2024.

\bibitem[Ren et~al.(2024)Ren, Xie, Mirzaei, Kreis, Liu, Torralba, Fidler, Kim, Ling, et~al.]{ren2024l4gm}
Jiawei Ren, Cheng Xie, Ashkan Mirzaei, Karsten Kreis, Ziwei Liu, Antonio Torralba, Sanja Fidler, Seung~Wook Kim, Huan Ling, et~al.
\newblock L4gm: Large 4d gaussian reconstruction model.
\newblock \emph{Advances in Neural Information Processing Systems}, 37:\penalty0 56828--56858, 2024.

\bibitem[Rombach et~al.(2022)Rombach, Blattmann, Lorenz, Esser, and Ommer]{rombach2022high}
Robin Rombach, Andreas Blattmann, Dominik Lorenz, Patrick Esser, and Bj{\"o}rn Ommer.
\newblock High-resolution image synthesis with latent diffusion models.
\newblock In \emph{Proceedings of the IEEE/CVF conference on computer vision and pattern recognition}, pages 10684--10695, 2022.

\bibitem[Ruiz et~al.(2023)Ruiz, Li, Jampani, Pritch, Rubinstein, and Aberman]{ruiz2023dreambooth}
Nataniel Ruiz, Yuanzhen Li, Varun Jampani, Yael Pritch, Michael Rubinstein, and Kfir Aberman.
\newblock Dreambooth: Fine tuning text-to-image diffusion models for subject-driven generation.
\newblock In \emph{Proceedings of the IEEE/CVF conference on computer vision and pattern recognition}, pages 22500--22510, 2023.

\bibitem[Shao et~al.(2024{\natexlab{a}})Shao, Pang, Zheng, Sun, and Liu]{shao2024human4dit}
Ruizhi Shao, Youxin Pang, Zerong Zheng, Jingxiang Sun, and Yebin Liu.
\newblock Human4dit: 360-degree human video generation with 4d diffusion transformer.
\newblock \emph{arXiv preprint arXiv:2405.17405}, 2024{\natexlab{a}}.

\bibitem[Shao et~al.(2024{\natexlab{b}})Shao, Wang, Li, Wang, Lin, Zhang, Fan, and Wang]{shao2024splattingavatar}
Zhijing Shao, Zhaolong Wang, Zhuang Li, Duotun Wang, Xiangru Lin, Yu Zhang, Mingming Fan, and Zeyu Wang.
\newblock Splattingavatar: Realistic real-time human avatars with mesh-embedded gaussian splatting.
\newblock In \emph{Proceedings of the IEEE/CVF Conference on Computer Vision and Pattern Recognition}, pages 1606--1616, 2024{\natexlab{b}}.

\bibitem[Su et~al.(2021)Su, Yu, Zollh{\"o}fer, and Rhodin]{su2021nerf}
Shih-Yang Su, Frank Yu, Michael Zollh{\"o}fer, and Helge Rhodin.
\newblock A-nerf: Articulated neural radiance fields for learning human shape, appearance, and pose.
\newblock \emph{Advances in neural information processing systems}, 34:\penalty0 12278--12291, 2021.

\bibitem[Su et~al.(2022)Su, Bagautdinov, and Rhodin]{su2022danbo}
Shih-Yang Su, Timur Bagautdinov, and Helge Rhodin.
\newblock Danbo: Disentangled articulated neural body representations via graph neural networks.
\newblock In \emph{European Conference on Computer Vision}, pages 107--124. Springer, 2022.

\bibitem[Voleti et~al.(2024)Voleti, Yao, Boss, Letts, Pankratz, Tochilkin, Laforte, Rombach, and Jampani]{voleti2024sv3d}
Vikram Voleti, Chun-Han Yao, Mark Boss, Adam Letts, David Pankratz, Dmitry Tochilkin, Christian Laforte, Robin Rombach, and Varun Jampani.
\newblock Sv3d: Novel multi-view synthesis and 3d generation from a single image using latent video diffusion.
\newblock In \emph{European Conference on Computer Vision}, pages 439--457. Springer, 2024.

\bibitem[Wang et~al.(2024{\natexlab{a}})Wang, Liu, Dou, Yu, Liang, Lin, Xie, Song, Li, and Wang]{wang2024disentangled}
Jionghao Wang, Yuan Liu, Zhiyang Dou, Zhengming Yu, Yongqing Liang, Cheng Lin, Rong Xie, Li Song, Xin Li, and Wenping Wang.
\newblock Disentangled clothed avatar generation from text descriptions.
\newblock In \emph{European Conference on Computer Vision}, pages 381--401. Springer, 2024{\natexlab{a}}.

\bibitem[Wang et~al.(2022)Wang, Schwarz, Geiger, and Tang]{wang2022arah}
Shaofei Wang, Katja Schwarz, Andreas Geiger, and Siyu Tang.
\newblock Arah: Animatable volume rendering of articulated human sdfs.
\newblock In \emph{European conference on computer vision}, pages 1--19. Springer, 2022.

\bibitem[Wang et~al.(2024{\natexlab{b}})Wang, Li, Lin, Zhai, Lin, Yang, Zhang, Liu, and Wang]{wang2024disco}
Tan Wang, Linjie Li, Kevin Lin, Yuanhao Zhai, Chung-Ching Lin, Zhengyuan Yang, Hanwang Zhang, Zicheng Liu, and Lijuan Wang.
\newblock Disco: Disentangled control for realistic human dance generation.
\newblock In \emph{Proceedings of the IEEE/CVF Conference on Computer Vision and Pattern Recognition}, pages 9326--9336, 2024{\natexlab{b}}.

\bibitem[Wang et~al.(2023{\natexlab{a}})Wang, Yuan, Zhang, Chen, Wang, Zhang, Shen, Zhao, and Zhou]{wang2023videocomposer}
Xiang Wang, Hangjie Yuan, Shiwei Zhang, Dayou Chen, Jiuniu Wang, Yingya Zhang, Yujun Shen, Deli Zhao, and Jingren Zhou.
\newblock Videocomposer: Compositional video synthesis with motion controllability.
\newblock \emph{Advances in Neural Information Processing Systems}, 36:\penalty0 7594--7611, 2023{\natexlab{a}}.

\bibitem[Wang et~al.(2024{\natexlab{c}})Wang, Zhang, Luo, Sun, Cui, Wang, Zhang, Wang, Li, Yu, et~al.]{wang2024emu3}
Xinlong Wang, Xiaosong Zhang, Zhengxiong Luo, Quan Sun, Yufeng Cui, Jinsheng Wang, Fan Zhang, Yueze Wang, Zhen Li, Qiying Yu, et~al.
\newblock Emu3: Next-token prediction is all you need.
\newblock \emph{arXiv preprint arXiv:2409.18869}, 2024{\natexlab{c}}.

\bibitem[Wang et~al.(2024{\natexlab{d}})Wang, Ma, Shao, Feng, Lai, and Li]{wang2024humancoser}
Yi Wang, Jian Ma, Ruizhi Shao, Qiao Feng, Yu-Kun Lai, and Kun Li.
\newblock Humancoser: Layered 3d human generation via semantic-aware diffusion model.
\newblock In \emph{2024 IEEE International Symposium on Mixed and Augmented Reality (ISMAR)}, pages 436--445. IEEE, 2024{\natexlab{d}}.

\bibitem[Wang et~al.(2024{\natexlab{e}})Wang, Wang, Liu, and Daniilidis]{wang2024tram}
Yufu Wang, Ziyun Wang, Lingjie Liu, and Kostas Daniilidis.
\newblock Tram: Global trajectory and motion of 3d humans from in-the-wild videos.
\newblock In \emph{European Conference on Computer Vision}, pages 467--487. Springer, 2024{\natexlab{e}}.

\bibitem[Wang et~al.(2004)Wang, Bovik, Sheikh, and Simoncelli]{wang2004image}
Zhou Wang, Alan~C Bovik, Hamid~R Sheikh, and Eero~P Simoncelli.
\newblock Image quality assessment: from error visibility to structural similarity.
\newblock \emph{IEEE transactions on image processing}, 13\penalty0 (4):\penalty0 600--612, 2004.

\bibitem[Wang et~al.(2023{\natexlab{b}})Wang, Lu, Wang, Bao, Li, Su, and Zhu]{wang2023prolificdreamer}
Zhengyi Wang, Cheng Lu, Yikai Wang, Fan Bao, Chongxuan Li, Hang Su, and Jun Zhu.
\newblock Prolificdreamer: High-fidelity and diverse text-to-3d generation with variational score distillation.
\newblock \emph{Advances in Neural Information Processing Systems}, 36:\penalty0 8406--8441, 2023{\natexlab{b}}.

\bibitem[Wang et~al.(2025)Wang, Dou, Liu, Lin, Dong, Guo, Zhang, Li, Wang, and Guo]{wang2025wonderhuman}
Zilong Wang, Zhiyang Dou, Yuan Liu, Cheng Lin, Xiao Dong, Yunhui Guo, Chenxu Zhang, Xin Li, Wenping Wang, and Xiaohu Guo.
\newblock Wonderhuman: Hallucinating unseen parts in dynamic 3d human reconstruction.
\newblock \emph{arXiv preprint arXiv:2502.01045}, 2025.

\bibitem[Wen et~al.(2024)Wen, Zhao, Ren, Schwing, and Wang]{wen2024gomavatar}
Jing Wen, Xiaoming Zhao, Zhongzheng Ren, Alexander~G Schwing, and Shenlong Wang.
\newblock Gomavatar: Efficient animatable human modeling from monocular video using gaussians-on-mesh.
\newblock In \emph{Proceedings of the IEEE/CVF Conference on Computer Vision and Pattern Recognition}, pages 2059--2069, 2024.

\bibitem[Weng et~al.(2022)Weng, Curless, Srinivasan, Barron, and Kemelmacher-Shlizerman]{weng2022humannerf}
Chung-Yi Weng, Brian Curless, Pratul~P Srinivasan, Jonathan~T Barron, and Ira Kemelmacher-Shlizerman.
\newblock Humannerf: Free-viewpoint rendering of moving people from monocular video.
\newblock In \emph{Proceedings of the IEEE/CVF conference on computer vision and pattern Recognition}, pages 16210--16220, 2022.

\bibitem[Xie et~al.(2024{\natexlab{a}})Xie, Mao, Bai, Zhang, Wang, Lin, Gu, Chen, Yang, and Shou]{xie2024showo}
Jinheng Xie, Weijia Mao, Zechen Bai, David~Junhao Zhang, Weihao Wang, Kevin~Qinghong Lin, Yuchao Gu, Zhijie Chen, Zhenheng Yang, and Mike~Zheng Shou.
\newblock Show-o: One single transformer to unify multimodal understanding and generation.
\newblock \emph{arXiv preprint arXiv:2408.12528}, 2024{\natexlab{a}}.

\bibitem[Xie et~al.(2024{\natexlab{b}})Xie, Yao, Voleti, Jiang, and Jampani]{xie2024sv4d}
Yiming Xie, Chun-Han Yao, Vikram Voleti, Huaizu Jiang, and Varun Jampani.
\newblock Sv4d: Dynamic 3d content generation with multi-frame and multi-view consistency.
\newblock \emph{arXiv preprint arXiv:2407.17470}, 2024{\natexlab{b}}.

\bibitem[Xu et~al.(2023)Xu, Yang, and Yang]{xu2023seeavatar}
Yuanyou Xu, Zongxin Yang, and Yi Yang.
\newblock Seeavatar: Photorealistic text-to-3d avatar generation with constrained geometry and appearance.
\newblock \emph{arXiv preprint arXiv:2312.08889}, 2023.

\bibitem[Xu et~al.(2024)Xu, Zhang, Liew, Yan, Liu, Zhang, Feng, and Shou]{xu2024magicanimate}
Zhongcong Xu, Jianfeng Zhang, Jun~Hao Liew, Hanshu Yan, Jia-Wei Liu, Chenxu Zhang, Jiashi Feng, and Mike~Zheng Shou.
\newblock Magicanimate: Temporally consistent human image animation using diffusion model.
\newblock In \emph{Proceedings of the IEEE/CVF Conference on Computer Vision and Pattern Recognition}, pages 1481--1490, 2024.

\bibitem[Yang et~al.(2024)Yang, Pan, Gu, and Zhang]{yang2024diffusion}
Zeyu Yang, Zijie Pan, Chun Gu, and Li Zhang.
\newblock Diffusion\textsuperscript{2}: Dynamic 3d content generation via score composition of video and multi-view diffusion models.
\newblock \emph{arXiv preprint arXiv:2404.02148}, 2024.

\bibitem[Ye et~al.(2023)Ye, Zhang, Liu, Han, and Yang]{ye2023ip}
Hu Ye, Jun Zhang, Sibo Liu, Xiao Han, and Wei Yang.
\newblock Ip-adapter: Text compatible image prompt adapter for text-to-image diffusion models.
\newblock \emph{arXiv preprint arXiv:2308.06721}, 2023.

\bibitem[Yu et~al.(2023)Yu, Cheng, Liu, Wu, and Lin]{yu2023monohuman}
Zhengming Yu, Wei Cheng, Xian Liu, Wayne Wu, and Kwan-Yee Lin.
\newblock Monohuman: Animatable human neural field from monocular video.
\newblock In \emph{Proceedings of the IEEE/CVF Conference on Computer Vision and Pattern Recognition}, pages 16943--16953, 2023.

\bibitem[Yu et~al.(2024)Yu, Chen, Huang, Sattler, and Geiger]{Yu2023MipSplatting}
Zehao Yu, Anpei Chen, Binbin Huang, Torsten Sattler, and Andreas Geiger.
\newblock Mip-splatting: Alias-free 3d gaussian splatting.
\newblock \emph{Conference on Computer Vision and Pattern Recognition (CVPR)}, 2024.

\bibitem[Zeng et~al.(2023)Zeng, Lu, Ji, Yao, Zhu, and Cao]{zeng2023avatarbooth}
Yifei Zeng, Yuanxun Lu, Xinya Ji, Yao Yao, Hao Zhu, and Xun Cao.
\newblock Avatarbooth: High-quality and customizable 3d human avatar generation.
\newblock \emph{arXiv preprint arXiv:2306.09864}, 2023.

\bibitem[Zhang et~al.(2025)Zhang, Xu, Wang, Yang, Zhao, Chen, and Guo]{zhang2025gaussian}
Bowen Zhang, Sicheng Xu, Chuxin Wang, Jiaolong Yang, Feng Zhao, Dong Chen, and Baining Guo.
\newblock Gaussian variation field diffusion for high-fidelity video-to-4d synthesis.
\newblock \emph{arXiv preprint arXiv:2507.23785}, 2025.

\bibitem[Zhang et~al.(2024{\natexlab{a}})Zhang, Chen, Yang, Qu, Wang, Chen, Long, Zhu, Du, and Zheng]{zhang2024avatarverse}
Huichao Zhang, Bowen Chen, Hao Yang, Liao Qu, Xu Wang, Li Chen, Chao Long, Feida Zhu, Daniel Du, and Min Zheng.
\newblock Avatarverse: High-quality \& stable 3d avatar creation from text and pose.
\newblock In \emph{Proceedings of the AAAI Conference on Artificial Intelligence}, pages 7124--7132, 2024{\natexlab{a}}.

\bibitem[Zhang et~al.(2024{\natexlab{b}})Zhang, Chen, Wang, Liu, Wang, and Qiao]{zhang20244diffusion}
Haiyu Zhang, Xinyuan Chen, Yaohui Wang, Xihui Liu, Yunhong Wang, and Yu Qiao.
\newblock 4diffusion: Multi-view video diffusion model for 4d generation.
\newblock \emph{Advances in Neural Information Processing Systems}, 37:\penalty0 15272--15295, 2024{\natexlab{b}}.

\bibitem[Zhang et~al.(2023)Zhang, Rao, and Agrawala]{zhang2023adding}
Lvmin Zhang, Anyi Rao, and Maneesh Agrawala.
\newblock Adding conditional control to text-to-image diffusion models.
\newblock In \emph{Proceedings of the IEEE/CVF international conference on computer vision}, pages 3836--3847, 2023.

\bibitem[Zhang et~al.(2018)Zhang, Isola, Efros, Shechtman, and Wang]{zhang2018unreasonable}
Richard Zhang, Phillip Isola, Alexei~A Efros, Eli Shechtman, and Oliver Wang.
\newblock The unreasonable effectiveness of deep features as a perceptual metric.
\newblock In \emph{Proceedings of the IEEE conference on computer vision and pattern recognition}, pages 586--595, 2018.

\bibitem[Zheng et~al.(2022)Zheng, Huang, Yu, Zhang, Guo, and Liu]{zheng2022structured}
Zerong Zheng, Han Huang, Tao Yu, Hongwen Zhang, Yandong Guo, and Yebin Liu.
\newblock Structured local radiance fields for human avatar modeling.
\newblock In \emph{Proceedings of the IEEE/CVF Conference on Computer Vision and Pattern Recognition}, pages 15893--15903, 2022.

\bibitem[Zheng et~al.(2023)Zheng, Zhao, Zhang, Liu, and Liu]{zheng2023avatarrex}
Zerong Zheng, Xiaochen Zhao, Hongwen Zhang, Boning Liu, and Yebin Liu.
\newblock Avatarrex: Real-time expressive full-body avatars.
\newblock \emph{ACM Transactions on Graphics (TOG)}, 42\penalty0 (4):\penalty0 1--19, 2023.

\bibitem[Zhu et~al.(2024)Zhu, Chen, Dai, Xu, Cao, Yao, Zhu, and Zhu]{zhu2024champ}
Shenhao Zhu, Junming~Leo Chen, Zuozhuo Dai, Yinghui Xu, Xun Cao, Yao Yao, Hao Zhu, and Siyu Zhu.
\newblock Champ: Controllable and consistent human image animation with 3d parametric guidance.
\newblock In \emph{European Conference on Computer Vision (ECCV)}, 2024.

\bibitem[Zielonka et~al.(2023)Zielonka, Bagautdinov, Saito, Zollh{\"o}fer, Thies, and Romero]{zielonka2023drivable}
Wojciech Zielonka, Timur Bagautdinov, Shunsuke Saito, Michael Zollh{\"o}fer, Justus Thies, and Javier Romero.
\newblock Drivable 3d gaussian avatars.
\newblock \emph{arXiv preprint arXiv:2311.08581}, 2023.

\end{thebibliography}
}
\clearpage

\appendix
\setcounter{table}{0}
\setcounter{figure}{0}
\renewcommand{\thetable}{\Alph{table}}
\renewcommand{\thefigure}{\Alph{figure}}

\maketitlesupplementary


\section{Implementation Details}

The avatar is finally optimized by:
\begin{equation}
    \small
    \mathcal{L} = \mathcal{L}_{\mathrm{rgb}} + \lambda_{\mathrm{LPIPS}}\mathcal{L}_{\mathrm{LPIPS}} + \lambda_{\mathrm{offset}}\mathcal{L}_{\mathrm{offset}} + \lambda_{\mathrm{scale}}\mathcal{L}_{\mathrm{scale}},
\end{equation}
where $\mathcal{L}_{\mathrm{rgb}}$ is the L1 photometric loss between the rendered image and the ground truth image, $\mathcal{L}_{\mathrm{LPIPS}}$ is the perceptual loss~\cite{zhang2018unreasonable}, and $\mathcal{L}_{\mathrm{offset}}, \mathcal{L}_{\mathrm{scale}}$ are regularization terms for the Gaussian points:
\begin{equation}
    \mathcal{L}_{\mathrm{offset}} = \frac{1}{N}\sum_{n=1}^N\|\Delta\boldsymbol{\mu}_n\|_2,
\end{equation}
\begin{equation}
    \mathcal{L}_{\mathrm{scale}} = \frac{1}{3}\sum_{c\in\{x,y,z\}}\left|\mathbf{s}_c - \frac{1}{3}\sum_{c\in\{x,y,z\}}\mathbf{s}_c\right|,
\end{equation}
encouraging the point cloud close to the body surface and keeping the Gaussians from being spiky.

During training, the ID finetuning and Super-resolution back-view generation are conducted on a single NVIDIA A800 with 80GB GPU memory. The identity dynamics injection requires approximately 20k steps and 20 hours, followed by an additional 6 hours to generate the double-resolution video. After the data preparation, the avatar model is trained on a single NVIDIA RTX 4090 GPU. It takes around 50k steps and 4 hours to converge. The reconstructed avatar could be rendered at 8$\sim$10 FPS.

\section{Evaluation on the Generated Video}

In this section, we present an evaluation of our generated videos. Table \ref{tab-fid} reports the quantitative results on the THuman4.0 dataset, following the format of Table \ref{tab-main} in the main text.
Since novel views are inherently unknown for our task, we additionally compute the Fréchet Inception Distance (FID)~\cite{heusel2017gans} to assess generative fidelity, which serves as a more suitable metric for novel view synthesis. Similar to LPIPS, the FID value is calculated within the minimal square bounding box enclosing the human region.

As shown in the top panel of the table, the resolution gap between the generative model and the 3D Gaussian model ($768$ v.s. $1150$) prevents the former from achieving competitive reconstruction quality with the latter. Supervising using the generated video leads to a slight degradation in input-view fidelity.
However, for novel views, the generated video exhibits higher quality compared to previous avatar-based approaches. Notably, the evaluated novel views differ from the “back-view” employed in our method, and generating videos for these views incurs substantially higher computational cost than directly rendering our reconstructed 3DGS. These quantitative results confirm that our back-view supervision strategy effectively distills the generative priors into the avatar model, achieving significant improvements over previous state-of-the-art methods.

\begin{table}[t]
\caption{\textbf{Quantitative comparisons} with AnimatableGaussians (AG for short)~\cite{li2024animatable} and the generated video (Gen. for short ) from finetuned Human4DiT on THuman4.0. The best and the second best are highlighted in \textbf{bold} and \underline{underlined} fonts, respectively.}
\label{tab-fid}
\centering

\begin{tabular}{@{}cccccc@{}}
\toprule
View              & Method & PSNR$\uparrow$ & SSIM$\uparrow$ & LPIPS$\downarrow$ & FID$\downarrow$ \\ \midrule
\multirow{3}{*}{\rotatebox{90}{Input}} & AG     & $\mathbf{34.06}$          & $\mathbf{0.9798}$         & $\mathbf{0.0554}$            & $\mathbf{17.91}$           \\
                  & Ours   & $\underline{32.97}$          & $\underline{0.9769}$         & $\underline{0.0558}$            & $\underline{19.16}$           \\
                  & Gen.   & $32.38$          & $0.9697$         & $0.0808$            & $38.55$           \\ \midrule
\multirow{3}{*}{\rotatebox{90}{Novel}} & AG     & $26.10$          & $0.9565$         & $0.1132$            & 100.34          \\
                  & Ours   & $\mathbf{26.39}$          & $\underline{0.9583}$         & $\underline{0.1053}$            & $\underline{71.06}$           \\
                  & Gen.   & $\underline{26.18}$          & $\mathbf{0.9598}$         & $\mathbf{0.1017}$            & $\mathbf{59.25}$           \\ \bottomrule
\end{tabular}
\end{table}

\section{More results}

Due to the page limit of the main paper, we show more results in the supplementary materials.
As shown in the Fig~\ref{fig:example}, our method achieves high-fidelity human reconstruction with dynamic details and supports dynamically consistent image synthesis at novel views.
To overcome the challenges for generating view-consistent and detailed results, we propose to leverage the advanced video generative model, Human4DiT, to generate the human motions from alternative perspective as an additional supervision signal. 
In addition, to solve the problem of insufficient resolution and inconsistent identity of Human4DiT, we inject the physical identity into the model through video fine-tuning and employ a patch-based denoising algorithm.
Quantitative and qualitative experiments prove the effectiveness of the proposed method.

To better compare the experimental results, the quantitative comparisons for each subject are shown in Tab. \ref{tab:detail-main}.

\begin{figure*}
    \centering
    \includegraphics[width=0.8\linewidth]{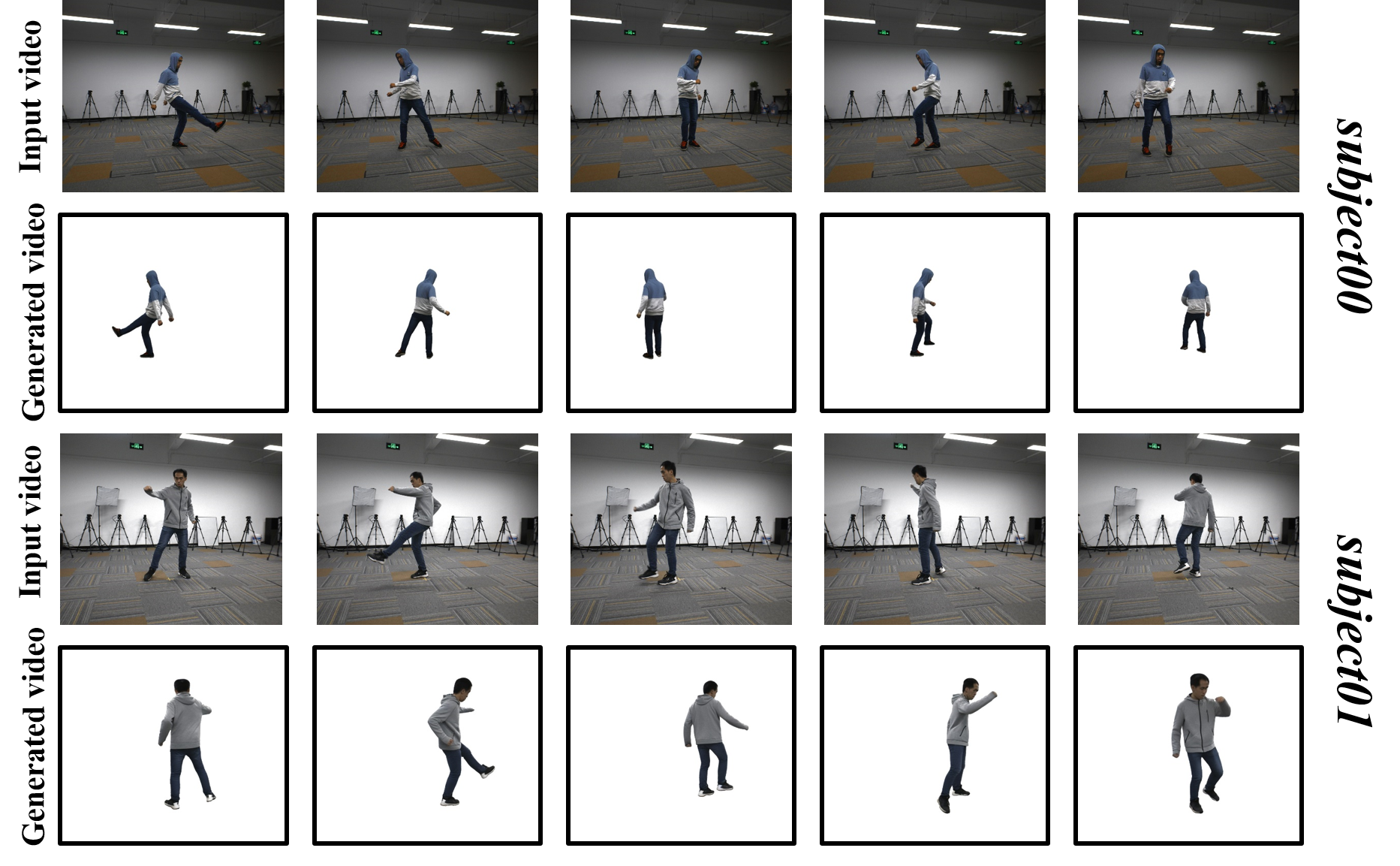}
    \caption{\textbf{Examples of input videos and generated videos.}}
    \label{fig:example}
\end{figure*}

\begin{table*}[t]
\caption{Quantitative comparisons on input view reconstruction and novel view synthesis with HumanNeRF~\cite{weng2022humannerf}, GaussianAvatar~\cite{hu2024gaussianavatar} and AnimatableGaussians (AG for short)~\cite{li2024animatable}. The best result is highlighted in \textbf{bold} fonts.}
\label{tab:detail-main}
\centering
\begin{tabular}{@{}ccccccccccc@{}}
\toprule
\multirow{2}{*}{Dataset}   & \multirow{2}{*}{Method}  & \multicolumn{3}{c}{Input View} & \multicolumn{3}{c}{Novel View 1} & \multicolumn{3}{c}{Novel View 2} \\ \cmidrule(lr){3-5} \cmidrule(lr){6-8} \cmidrule(l){9-11} 
                           &                & PSNR$\uparrow$     & SSIM$\uparrow$     & LPIPS$\downarrow$    & PSNR$\uparrow$      & SSIM$\uparrow$      & LPIPS$\downarrow$    & PSNR$\uparrow$      & SSIM$\uparrow$      & LPIPS$\downarrow$    \\ \midrule
\multirow{4}{*}{\rotatebox{90}{\begin{tabular}[c]{@{}c@{}}\small \textit{THuman4.0}\\ \textit{subject00}\end{tabular}}} & HumanNeRF               & $30.84$    & $0.9597$   & $0.0940$    & $27.01$    & $0.9628$    & $0.1181$    & $26.41$    & $0.9586$    & $0.1229$    \\
                           & GaussianAvatar          & $31.08$    & $0.9744$   & $0.0777$   & $23.10$     & $0.9668$    & $0.1317$    & $22.94$    & $0.9641$    & $0.1345$    \\
                           & AG                      & $\mathbf{32.76}$       & $\mathbf{0.9820}$    & $\mathbf{0.0535}$    & $28.43$    & $0.9720$    & $0.1047$    & $27.83$    & $0.9690$    & $0.1130$    \\
                           & Ours                    & $32.32$    & $0.9801$   & $0.0532$   & $\mathbf{28.51}$    & $\mathbf{0.9738}$    & $\mathbf{0.0959}$    & $\mathbf{28.08}$    & $\mathbf{0.9710}$    & $\mathbf{0.1028}$    \\ \midrule
\multirow{4}{*}{\rotatebox{90}{\begin{tabular}[c]{@{}c@{}}\small \textit{THuman4.0}\\ \textit{subject01}\end{tabular}}} & HumanNeRF               & $30.22$    & $0.9393$   & $0.0883$   & $26.27$    & $0.9386$    & $\mathbf{0.0994}$    & $\mathbf{24.91}$    & $0.9186$    & $\mathbf{0.1168}$    \\
                           & GaussianAvatar          & $30.91$    & $0.9652$   & $0.0906$   & $22.50$     & $0.9500$      & $0.1286$    & $21.70$     & $0.9373$    & $0.1398$    \\
                           & AG                      & $\mathbf{34.19}$    & $\mathbf{0.9749}$   & $0.0623$   & $26.23$    & $0.9540$    & $0.1157$    & $24.41$    & $0.9388$    & $0.1360$    \\
                           & Ours                    & $32.77$    & $0.9749$   & $\mathbf{0.0609}$   & $\mathbf{26.39}$    & $\mathbf{0.9573}$     & $0.1032$    & $24.68$    & $\mathbf{0.9418}$    & $0.1243$    \\ \midrule
\multirow{4}{*}{\rotatebox{90}{\begin{tabular}[c]{@{}c@{}}\small \textit{THuman4.0}\\ \textit{subject02}\end{tabular}}} & HumanNeRF               & $31.00$       & $0.9556$   & $0.0699$   & $24.25$    & $0.9301$    & $\mathbf{0.0977}$    & $23.58$    & $0.9344$    & $0.1091$    \\
                           & GaussianAvatar          & $31.39$    & $0.9731$   & $0.0777$   & $21.16$    & $0.9430$     & $0.1211$    & $21.41$    & $0.9485$    & $0.1295$    \\
                           & AG                      & $\mathbf{35.24}$    & $\mathbf{0.9824}$   & $\mathbf{0.0503}$   & $25.00$    & $\mathbf{0.9505}$    & $0.1010$    & $24.71$    & $0.9549$    & $0.1088$    \\
                           & Ours                    & $33.82$    & $0.9788$   & $0.0533$   & $\mathbf{25.46}$    & $\mathbf{0.9505}$    & $0.0995$    & $\mathbf{25.18}$       & $\mathbf{0.9552}$    & $\mathbf{0.1061}$    \\ \midrule
\multirow{4}{*}{\rotatebox{90}{\begin{tabular}[c]{@{}c@{}}\small \textit{Mono2K}\\ \textit{male}\end{tabular}}} & HumanNeRF               & $26.90$     & $0.9206$   & $0.0836$   & $25.21$    & $0.9198$    & $0.0879$    & $25.27$    & $0.9244$    & $0.0948$    \\
                           & GaussianAvatar          & $28.35$    & $0.9693$   & $0.0758$   & $19.95$    & $0.9473$    & $0.1084$    & $20.07$    & $0.9506$    & $0.1144$    \\
                           & AG                      & $\mathbf{32.44}$    & $\mathbf{0.9774}$   & $\mathbf{0.0566}$   & $\mathbf{25.40}$     & $0.9601$    & $0.0841$    & $25.50$     & $0.9561$    & $0.0987$    \\
                           & Ours                    & $31.45$    & $0.9733$   & $0.0656$   & $25.28$    & $\mathbf{0.9627}$    & $\mathbf{0.0786}$    & $\mathbf{25.53}$    & $\mathbf{0.9623}$    & $\mathbf{0.0892}$    \\ \midrule
\multirow{4}{*}{\rotatebox{90}{\begin{tabular}[c]{@{}c@{}}\small \textit{Mono2K}\\ \textit{female}\end{tabular}}}          & HumanNeRF      & $28.91$    & $0.9424$   & $0.0808$   & $26.83$    & $0.9412$    & $0.0866$    & $25.62$    & $0.9282$    & $0.0933$    \\
                           & GaussianAvatar          & $29.91$    & $0.9764$   & $0.0709$   & $22.57$    & $0.9625$    & $0.1058$    & $21.58$    & $0.9578$    & $0.1098$    \\
                           & AG                      & $\mathbf{33.86}$    & $\mathbf{0.9839}$   & $\mathbf{0.0500}$     & $27.18$    & $0.9706$    & $0.0796$    & $25.70$     & $0.9621$    & $0.0957$    \\
                           & Ours                    & $32.94$     & $0.9803$   & $0.0572$   & $\mathbf{27.24}$    & $\mathbf{0.9713}$    & $\mathbf{0.0790}$     &    $\mathbf{25.94}$    & $\mathbf{0.9643}$    & $\mathbf{0.0904}$     \\ \bottomrule
\end{tabular}
\end{table*}

\section{Novel Pose Synthesis}
\begin{figure}[t]
    \centering
    \includegraphics[width=\linewidth]{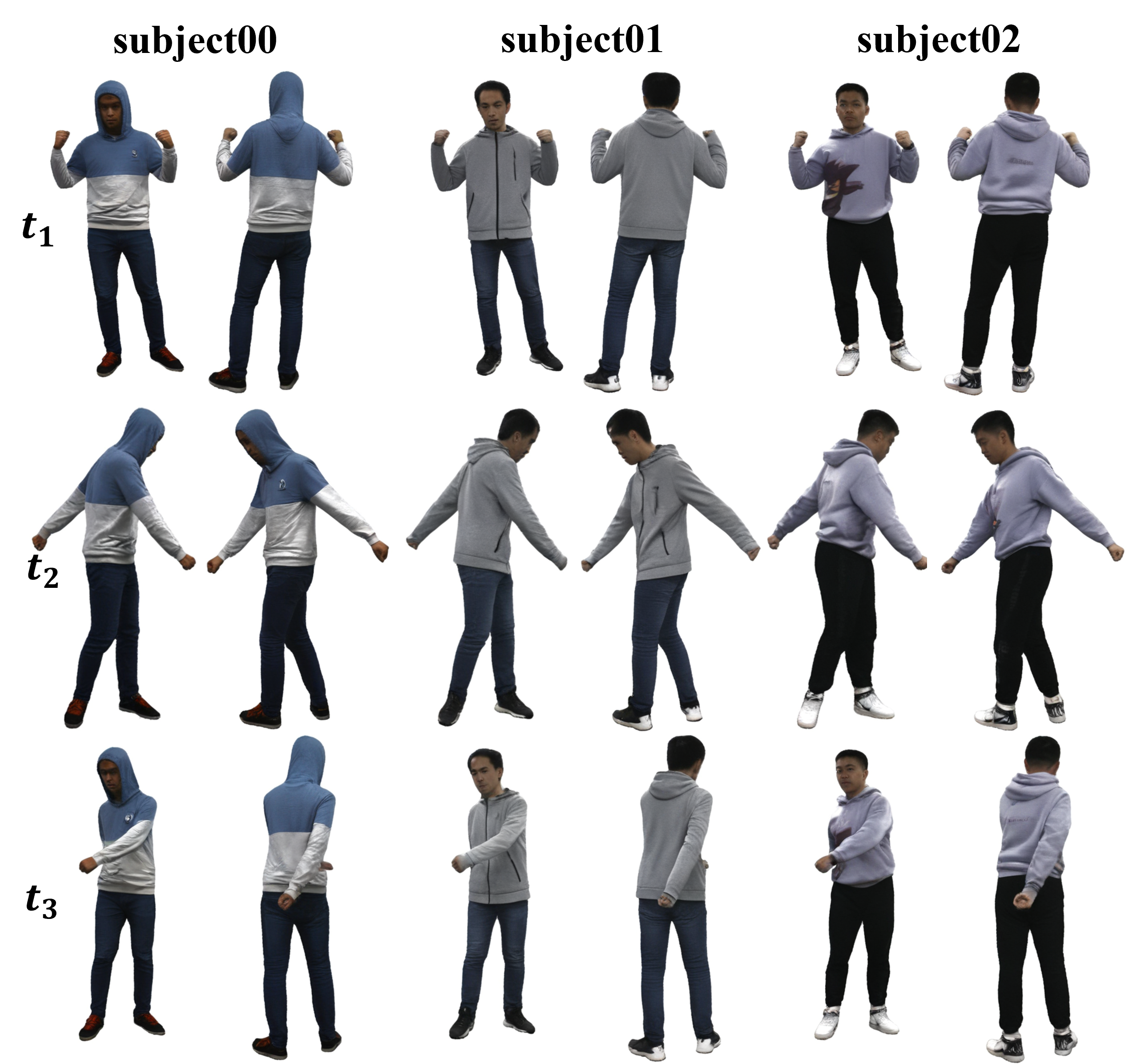}
    \caption{\textbf{Novel Pose Synthesis.} We synthesized all subjects in \textit{THuman4.0} under the same test pose sequence. Our method is able to generate rich dynamic details that remain consistent with those observed in the training frames.}
    \label{fig:novel}
\end{figure}

After constructing the human avatar, we can directly deform the learned model to animate the character under arbitrary poses. Furthermore, a similar reconstruction strategy can be employed to enable novel pose synthesis. Specifically, given the novel poses, we generate front-view and back-view videos simultaneously and incorporate them as additional training data for the learned avatar. This approach leverages both the inherent human structural priors and the generative priors, similar to our reconstruction process.

Next, we present several synthesized results obtained using this strategy. A pose sequence from \textit{THuman4.0-subject02}, which does not overlap with the training data, is selected as the test sequence. Figure~\ref{fig:novel} illustrates the synthesized results for all subjects in the \textit{THuman4.0} dataset, with each row corresponding to the poses at different timestamps. Our method is able to generate rich dynamic details that remain consistent with those observed in the training frames, which demonstrate that our Physical Identity Inversion strategy effectively captures the physical properties of the human body.

\section{In-the-wild Examples}
\begin{figure}[t]
    \centering
    \includegraphics[width=\linewidth]{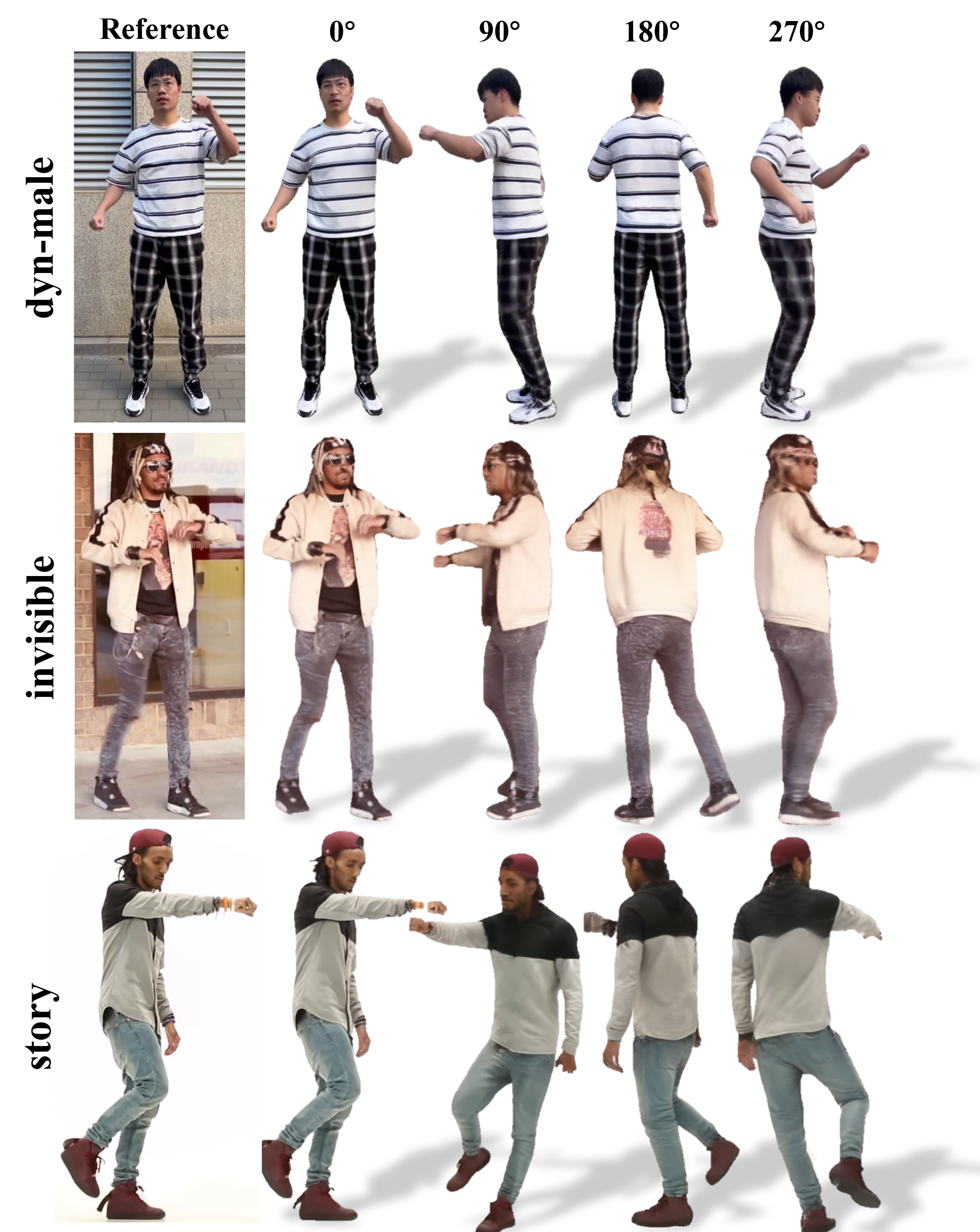}
    \caption{\textbf{In-the-wild examples by our method.} The first example is acquired from GaussianAavatar~\cite{hu2024gaussianavatar} while the rest two are downloaded from Youtube. Even without back-view observations, our method is capable of synthesizing a plausible full-body appearance.}
    \label{fig:wild}
\end{figure}
In the main text, we evaluate all methods using self-rotating sequences to ensure fair comparisons. However, this condition is not necessary for our approach. Figure~\ref{fig:wild} presents reconstructed results on in-the-wild examples with only front-view frames observed. The example in the top row is taken from GaussianAvatar~\cite{hu2024gaussianavatar}, while the remaining examples are sourced from YouTube videos that were also used in HumanNeRF~\cite{weng2022humannerf}. These examples are included solely for academic purposes. 
Prior to processing the in-the-wild data in our pipeline, the estimated poses are refined using the same strategy as the first-stage training of GaussianAvatar. As shown in the figure, even without back-view observations, our method is capable of synthesizing a plausible full-body appearance.

\section{Potential Social Impacts}

Our approach automates the creation of digital images representing any individual's identity from monocular video, where the video can be effortlessly captured using a smartphone. However, this feature also carries the risk of producing fabricated motion sequences that the person never executed, a concern that must be thoroughly addressed before deployment.

\end{document}